\documentclass[lettersize,journal]{IEEEtran}
\usepackage[colorlinks,
          linkcolor=blue,
          anchorcolor=green,
          urlcolor=blue,
          citecolor=blue,
          ]{hyperref}
\usepackage{adjustbox}
\ifCLASSOPTIONcompsoc
\usepackage[caption=false, font=normalsize, labelfont=sf, textfont=sf]{subfig}
\else
\usepackage[caption=false, font=footnotesize]{subfig}
\fi
\usepackage{url}
\usepackage[normalem]{ulem}

\usepackage[flushleft]{threeparttable}
\usepackage{booktabs}
\usepackage{makecell}
\usepackage{multirow}
\usepackage{graphics}
\usepackage{graphicx}
\usepackage{algorithm}
\usepackage{algpseudocode}
\usepackage{amsthm,amsmath,amssymb}

\usepackage{mathrsfs}
\usepackage{cleveref}
\usepackage{cite}

\usepackage{subfig}
\def\BibTeX{{\rm B\kern-.05em{\sc i\kern-.025em b}\kern-.08em
    T\kern-.1667em\lower.7ex\hbox{E}\kern-.125emX}}
\usepackage{balance}

\newcounter{subeq}

\begin{document}
\title{Towards Generalization-Oriented Models for Vehicle Routing Problems with Mixture-of-Experts}

\author{Chang-Hao Miao, Yun-Tian Zhang, Tong-Yu Wu,\\ Fang Deng, \emph{Fellow}, \emph{IEEE} and Chen Chen, \emph{Member}, \emph{IEEE}\thanks{This work was supported by the National Key Research and Development Program of China No.2022ZD0119703; in part by the National Natural Science Foundations of China (NSFC) under Grant 62273044 and 62022015; in part by the National Natural Science Foundation of China National Science Fund for Distinguished Young Scholars under Grant 62025301; in part by the National Natural Science Foundation of China Basic Science Center Program under Grant 62088101. \emph{(Corresponding authors: Chen Chen)}}
\thanks{Chang-Hao Miao, Yun-Tian Zhang, Tong-Yu Wu, Fang Deng, and Chen Chen are with the State Key Laboratory of Autonomous Intelligent Unmanned Systems, Beijing Institute of Technology, Beijing 100081, China (e-mail: xiaofan@bit.edu.cn). Yun-Tian Zhang is also with the Department of Computer, Control and Management Engineering ``Antonio Ruberti", Sapienza University of Rome, Rome, 00185, Italy. Fang Deng and Chen Chen are also with the School of AI, Beijing Institute of Technology, Beijing 100081, China.}}

\markboth{\parbox[t]{\textwidth}{This work has been submitted to the IEEE for possible publication.\newline Copyright may be transferred without notice, after which this version may no longer be accessible.}}{How to Use the IEEEtran \LaTeX \ Templates}

\maketitle

\begin{abstract}
In recent years, Deep Reinforcement Learning (DRL) has achieved substantial progress on Vehicle Routing Problems (VRPs). However, existing DRL-based methods are typically trained on instances generated from a uniform distribution, which limits their performance under real-world distribution shifts. In this paper, we aim to develop a generalization-oriented model that partitions the policy network into multiple modules and adaptively recombines modules to form specific policies during inference. Specifically, we propose Residual Refined Experts with Instance-level Gating (R2E-IG) to improve cross-distribution generalization. Our contributions are threefold: (1) We introduce a Residual Refined Expert (R2E) architecture that enhance expert expressiveness via residual refinement; (2) We design an instance-level gating mechanism that learns distribution-aware instance representations and routes inputs to suitable modules; (3) We propose a mixed-distribution training mechanism equipped with Dynamic Weight Adaption (DWA), which dynamically reweights training data from different distributions to emphasize more informative ones. Extensive experiments show that R2E-IG achieves competitive performance against state-of-the-art baselines on both in-distribution and out-of-distribution instances across synthetic and benchmark datasets. Moreover, R2E-IG is generic and can be easily integrated into existing DRL-based methods to further improve performance.
\end{abstract}


\section{Introduction}
\IEEEPARstart{V}{ehicle} Routing problems (VRPs) are classic Combinatorial Optimization Problems with broad real-world applications, including logistics \cite{konstantakopoulos2022vehicle}, industrial manufacturing \cite{zhang2023review}, and transportation management \cite{wu2023neural}, and have been extensively studied in operations research and machine learning for decades. Due to their Non-deterministic Polynomial Hard (NP-hard) nature, VRPs are computationally challenging, which limits the practicality of exact solvers \cite{bengio2021machine}. As an alternative, heuristic methods have been developed to produce satisfactory solutions within acceptable time. However, heuristic methods heavily rely on complex hand-crafted rules, leaving substantial room for further improvement. In recent years, Deep Reinforcement Learning (DRL) has been widely applied to solve VRPs and has achieved notable success. Compared to traditional methods, DRL-based methods learn efficient policies from large amounts of training data, enabling high-quality solutions with low computational cost. 

However, existing DRL-based methods are typically trained and tested under same instance distribution, which prevents the model from capturing the characteristics of diverse instance distributions. For example, many representative attention-based models (e.g., AM \cite{kool2018attention} and POMO \cite{kwon2020pomo}) are trained with the implicit assumption that nodes are uniformly distributed, whereas real-world instances (e.g., CVRPLIB \cite{uchoa2017new} and TSPLIB \cite{reinelt1991tsplib}) often deviate significantly from this assumption. As a result, although these methods achieve competitive results on In-Distribution (ID) instances, they exhibit poor cross-distribution generalization when facing unseen Out-of-Distribution (OoD) instances \cite{joshi2021learning}, which limits their practical applicability.

To alleviate the cross-distribution generalization issue, a number of preliminary attempts have been made, including multi-distribution training \cite{jiang2022learning,bi2022learning}, ensemble methods\cite{jiang2023ensemble,gao2024towards}, and meta-learning \cite{manchanda2022generalization,zhou2023towards}. The common idea is to expose the model to instances from multiple distributions during training, so that it can handle several representative instance distributions simultaneously. While intuitive and empirically effective, these approaches still have notable limitations. The methods can only generalize to the distributions seen during training, rather than truly learning and exploiting distribution characteristics. Meanwhile, these methods typically yield a single, fixed policy with a predetermined network structure. Such a rigid architecture lacks the flexibility to adaptively adjust the policy according to the distribution characteristics of the given instances, leaving substantial room for further improvement in cross-distribution generalization. Different from existing studies, we aim to develop a generalization-oriented model that partitions the policy network into multiple modules and adaptively recombines modules to form specific policies during inference.

Motivated by the recent advance of large language models (LLMs) \cite{dubey2024llama,achiam2023gpt,liu2024deepseek}, we propose Residual Refined Experts with Instance-level Gating (R2E-IG) for enhancing cross-distribution generalization. In Transformer architectures, a common way to scale model capacity is to replace the standard Feed-Forward Network (FFN) with a Mixture-of-Experts (MoE) layer that contains multiple experts, each with its own parameters. A lightweight gating network then assigns each input to a subset of experts, so that only the selected experts are activated. This design increases effective capacity without incurring a proportional rise in computation, making it attractive for large-scale training and deployment. As to cross-distribution generalization, we can leverage the dynamic activation property of MoE to flexibly compose an appropriate network structure based on the distribution characteristics of each instance. Meanwhile, since most existing DRL-based methods for VRPs are built upon Transformer architectures, our method can be integrated in a plug-and-play manner, demonstrating its generality.

Our contributions can be summarized as follows: (1) We introduce a Residual Refined Expert (R2E) architecture building upon vanilla expert, which improves expert expressiveness via residual refinement; (2) We design an instance-level gating mechanism that learns informative distribution-aware representations for each instance and uses them to activate the most suitable experts; (3) We propose a mixed-distribution training mechanism, equipped with a Dynamic Weight Adaptation (DWA) technique, to adaptively adjust the mixture proportions of training data across distributions during training, encouraging the model to focus on more informative distributions and improving training efficiency. Extensive experimental results demonstrate that R2E-IG outperforms state-of-the-art baselines in both ID and OoD instances, including synthetic and real-world datasets. Furthermore, we conduct extensive experiments to evaluate each proposed component and to analyze the effects of various MoE configurations.

The rest of the paper is organized as follows. Section \ref{RW} briefly reviews related work and presents our motivation. Section \ref{PS} briefly introduces the preliminaris and  Section \ref{Methodology} describes the proposed methodology. Section \ref{exp} reports the experimental results. Finally, Section \ref{Conclusion} concludes the paper and discusses future work.

\section{Background and Motivation}
\label{RW}

\noindent In this section, we review methodological developments in neural methods for VRPs and recent efforts on their generalization. We also briefly summarize relevant advances in MoE. Based on this review, we identify the limitations of existing methods and motivate our work.

\subsection{Neural Methods for Vehicle Routing Problems}
Most of recent methods can be divided into two categories: \emph{1) Learn-for-Construction:} the model learns to construct solutions sequentially without additional modifications or improvements. Vinyals et al. \cite{vinyals2015pointer} first proposed Pointer Network to solve routing problems in a supervised manner, inspiring a series of subsequent works based on reinforcement learning \cite{bello2017neural,nazari2018reinforcement}. Benefited from the structure of Transformer \cite{vaswani2017attention}, Kool et al. \cite{kool2018attention} proposed an attention model (AM) to tackle with routing problems. Kwon et al. \cite{kwon2020pomo} further proposed policy optimization with multiple optima, which enhanced AM by utilizing the symmetrical property of the problem. In addition, graph neural networks have also been used to solve routing problems \cite{khalil2017learning}, which requires to be combined with post-search to obtain feasible solutions (e.g. beam search \cite{joshi2019efficient}, Monte-Carlo tree search \cite{fu2021generalize}, and dynamic programming \cite{kool2022deep}). \emph{2) Learn-for-Improvement:} the model learns to improve an initial solution by heuristics until the termination conditions are satisfied. One of the most common methods is to combine neural networks with heuristic methods to guide the heuristic search process (e.g. large neighborhood search \cite{hottung2020neural}, guided local search \cite{hudson2021graph}, and Lin-Kernighan-Helsgaun heuristic \cite{xin2021neurolkh}). Considering the outstanding capabilities of various heuristic operators in routing problems, some researchers have tried to improve the performance by learning to guide the execution of heuristic operators (e.g. 2-opt \cite{d2020learning}, 3-opt \cite{sui2021learning}, k-opt \cite{ma2023learning}). In general, learn-for-improvement often achieves better performance than learn-for-construction at the expense of time exploration. Beyond VRPs, neural methods have also been applied to other COPs, such as the facility location problems \cite{miao2024deep}, job shop scheduling problems \cite{zhang2020learning}, and location routing problems \cite{miao2025end}.

\begin{figure*}[t]
\centering
\setlength{\tabcolsep}{1pt}
\begin{tabular}{ccccccc}
\hspace{-.4cm}
\subfloat[Uniform]{\includegraphics[width=.15\linewidth]{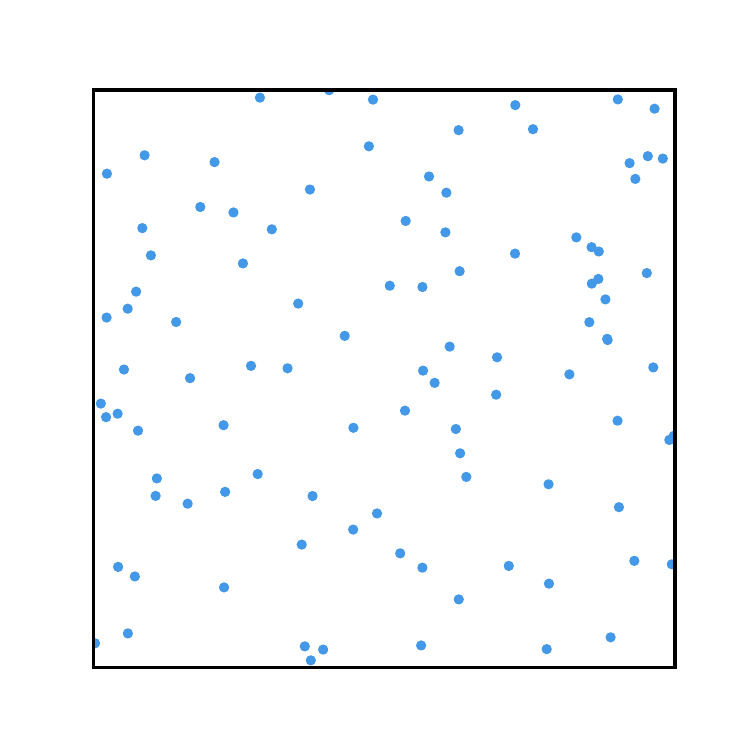}} &
\hspace{-.4cm}
\subfloat[Cluster]{\includegraphics[width=.15\linewidth]{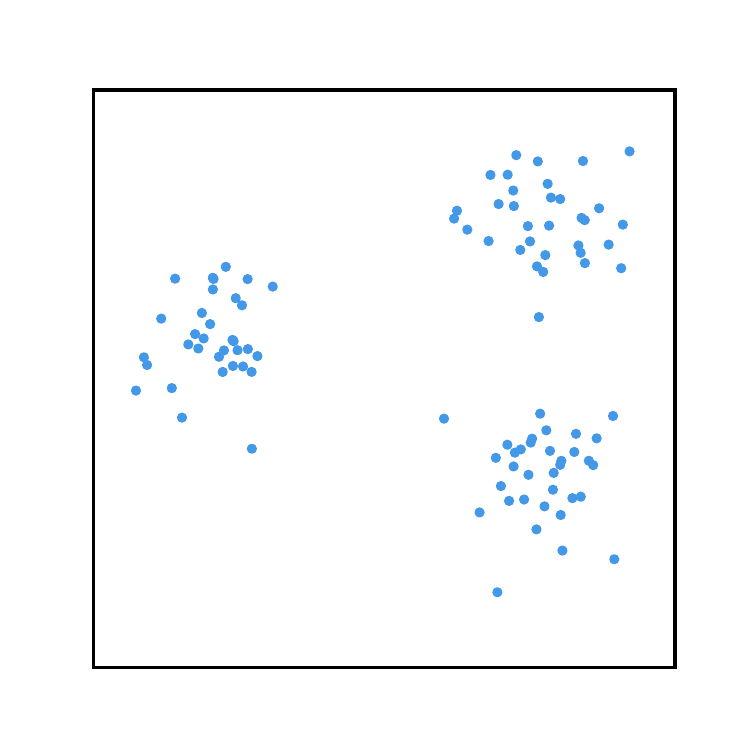}} &
\hspace{-.4cm}
\subfloat[Mixed]{\includegraphics[width=.15\linewidth]{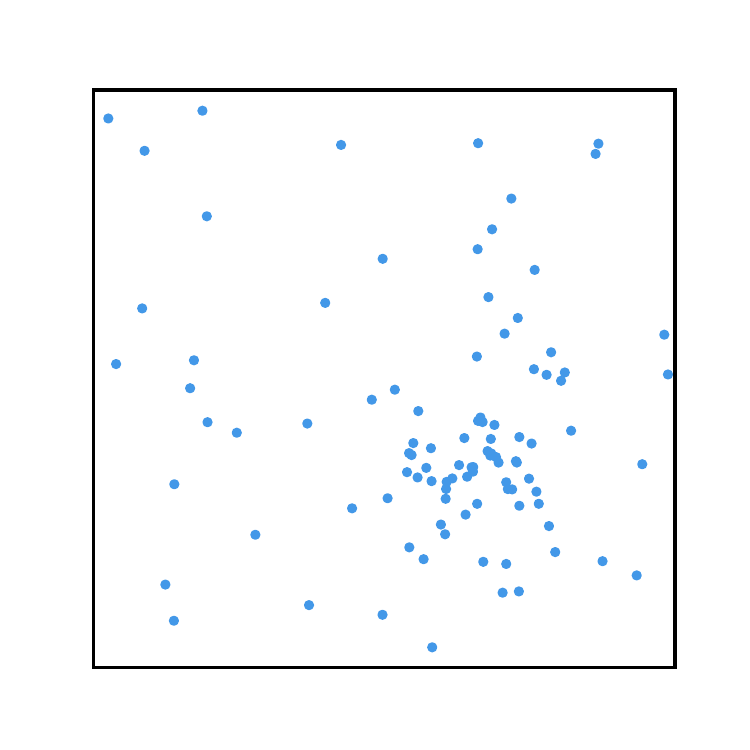}}& 
\hspace{-.4cm}
\subfloat[Explosion]{\includegraphics[width=.15\linewidth]{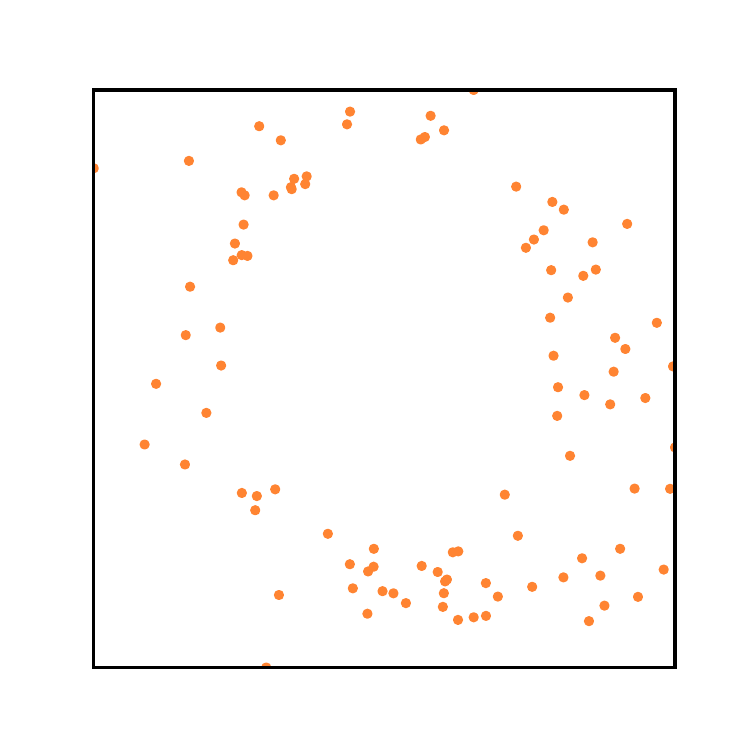}} &
\hspace{-.4cm}
\subfloat[Expansion]{\includegraphics[width=.15\linewidth]{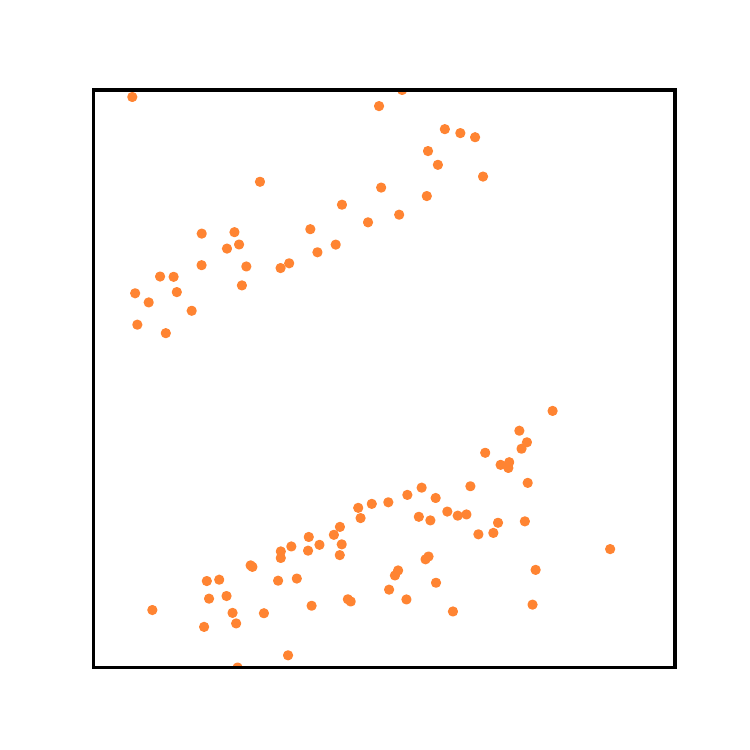}} &
\hspace{-.4cm}
\subfloat[Grid]{\includegraphics[width=.15\linewidth]{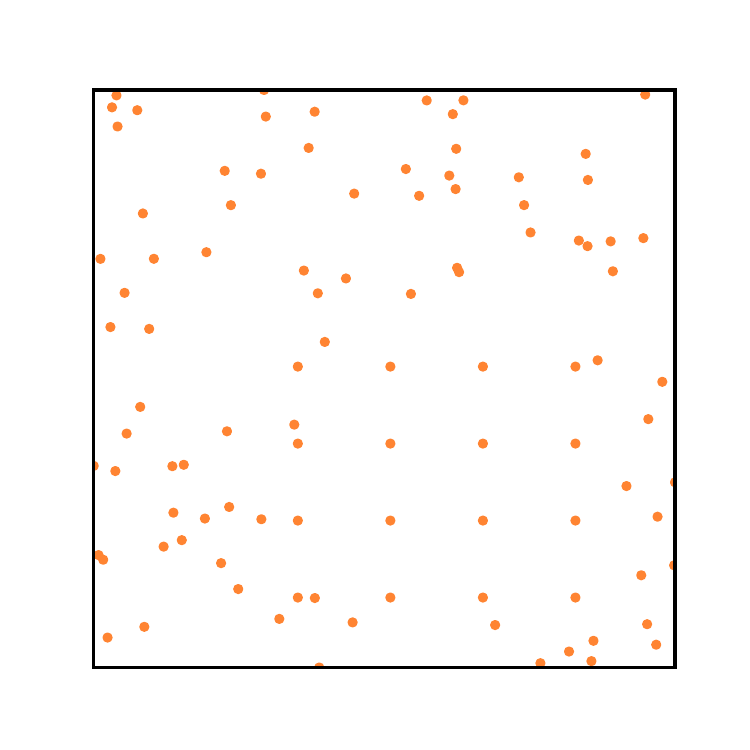}} &
\hspace{-.4cm}
\subfloat[Implosion]{\includegraphics[width=.15\linewidth]{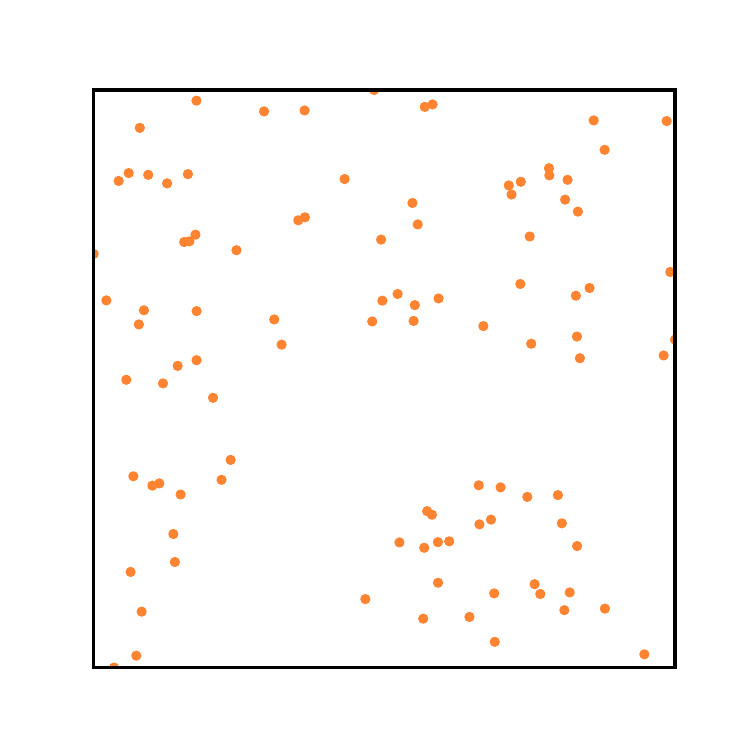}} \vspace{-.4cm} \\ 
\hspace{-.4cm}
\subfloat[X-n125-k30]{\includegraphics[width=.15\linewidth]{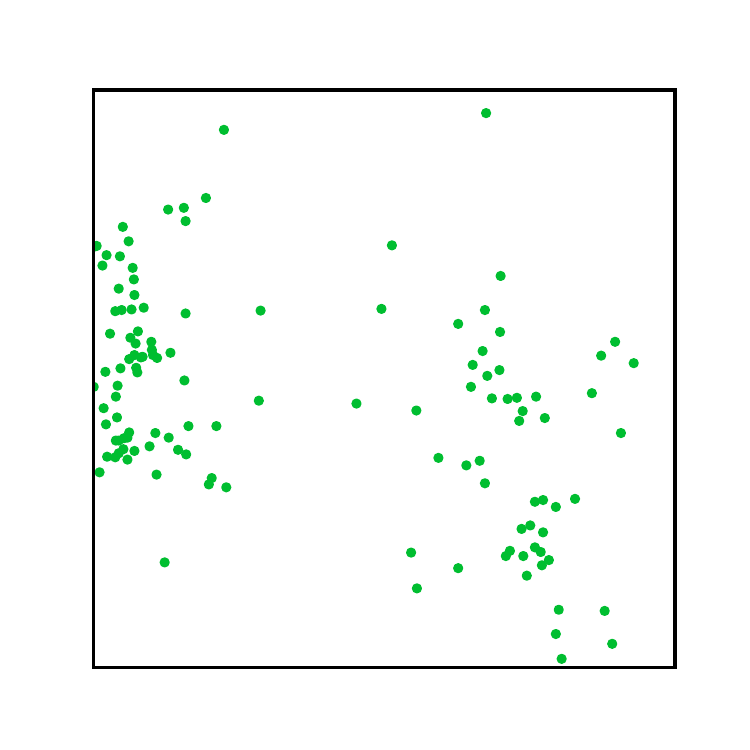}} &
\hspace{-.4cm}
\subfloat[X-n181-k23]{\includegraphics[width=.15\linewidth]{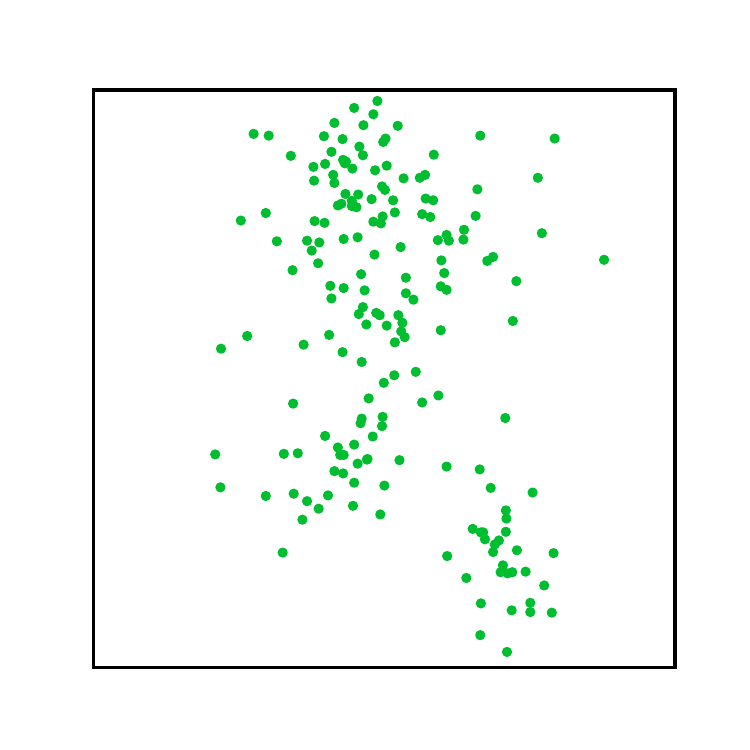}} &
\hspace{-.4cm}
\subfloat[X-n190-k7]{\includegraphics[width=.15\linewidth]{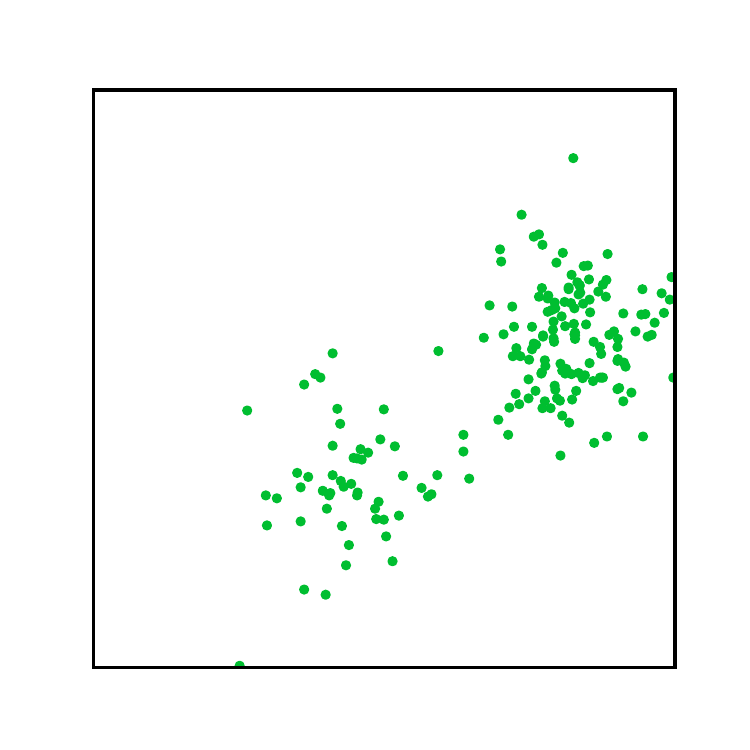}}& 
\hspace{-.4cm}
\subfloat[lin105]{\includegraphics[width=.15\linewidth]{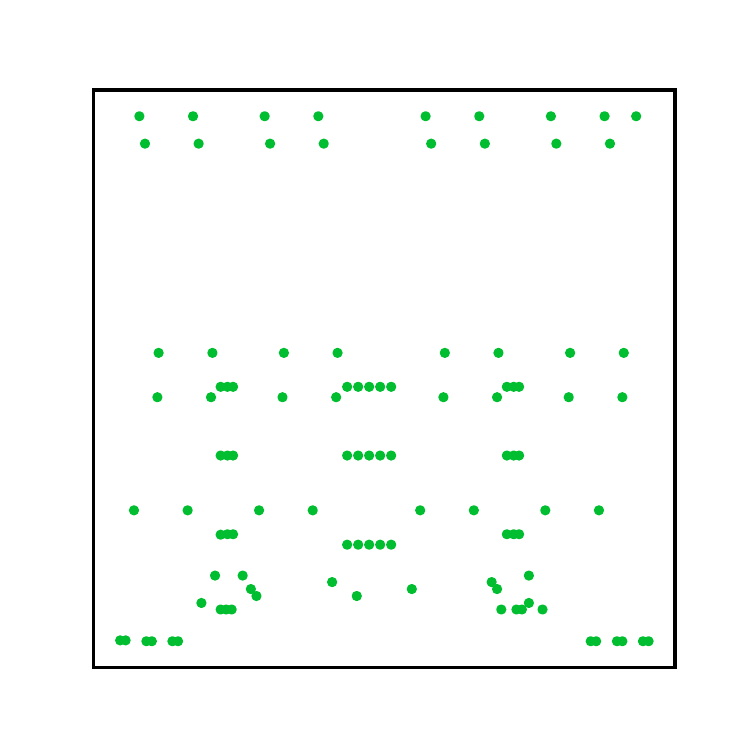}} &
\hspace{-.4cm}
\subfloat[bier127]{\includegraphics[width=.15\linewidth]{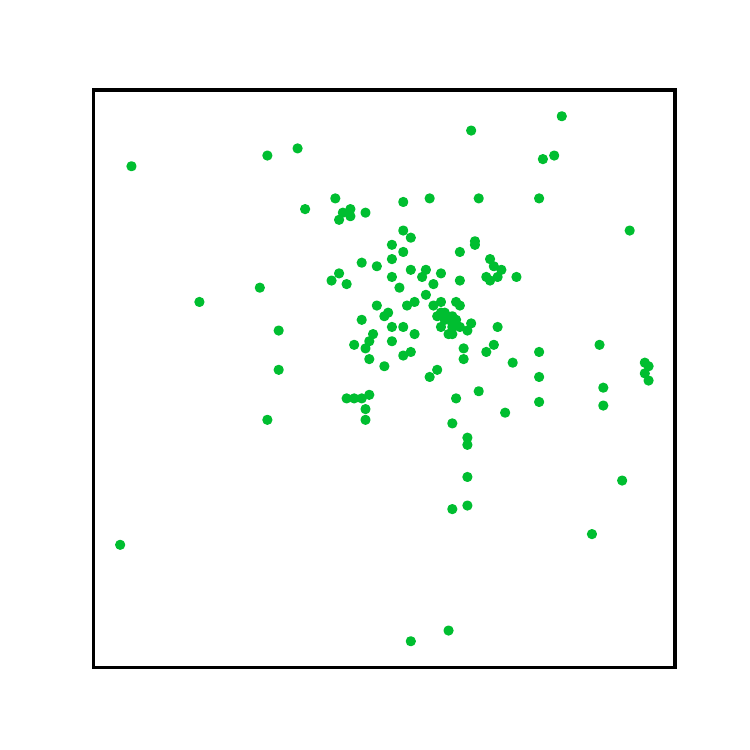}} &
\hspace{-.4cm}
\subfloat[pr144]{\includegraphics[width=.15\linewidth]{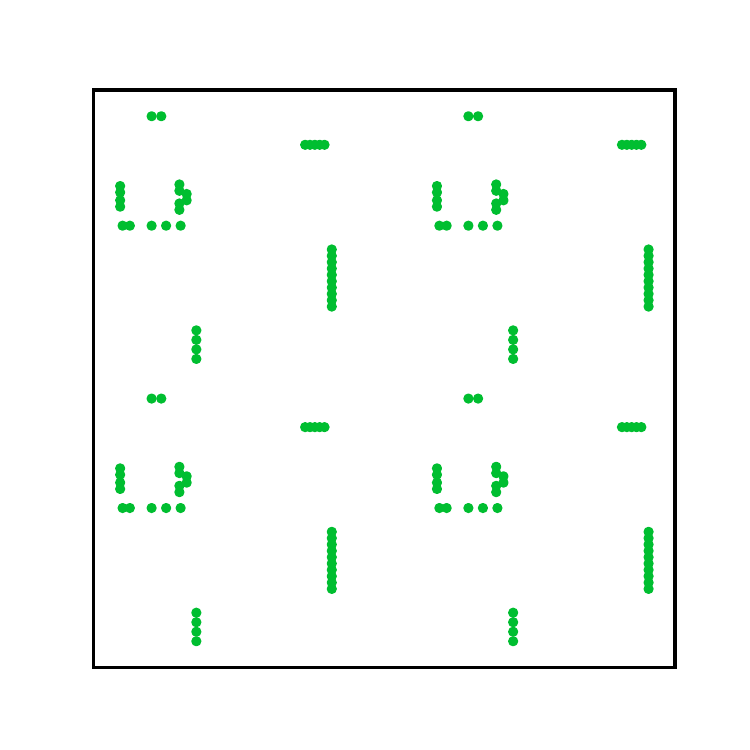}} &
\hspace{-.4cm}
\subfloat[u159]{\includegraphics[width=.15\linewidth]{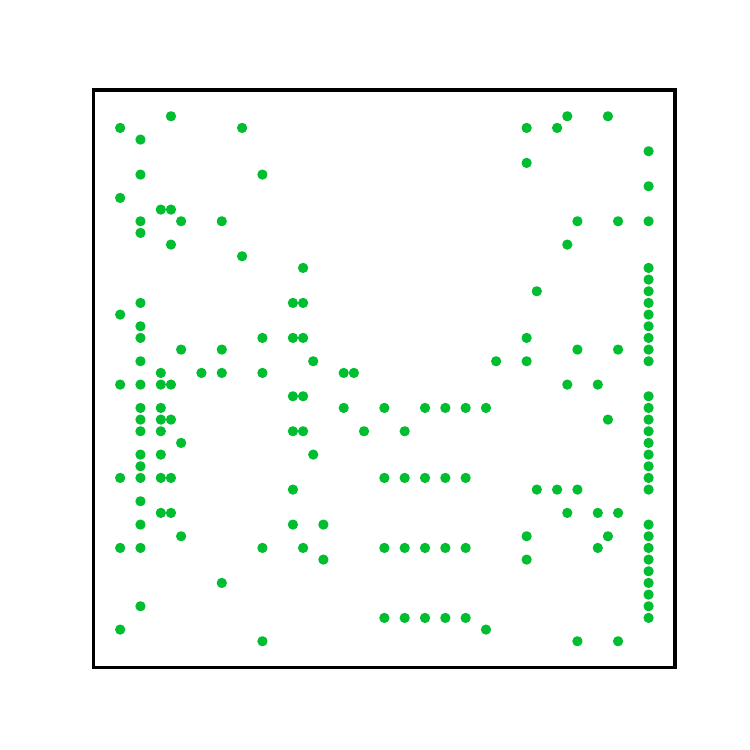}}
\end{tabular}
\caption{The VRP instances with various distributions in this paper, where (a)-(c) are ID instances, (d)-(g) are OoD instances, and (h)-(n) are drawn from benchmark datasets (e.g., CVRPLIB \cite{uchoa2017new} and TSPLIB \cite{reinelt1991tsplib}).}
\label{fig:distribution}
\end{figure*}

\subsection{Generalization of Neural Methods}
The above methods are trained and tested under the same instance distribution, which limits their ability to generalize across various distributions \cite{joshi2021learning}. To improve the cross-distribution generalization of neural methods, numerous preliminary attempts have explored training on instances drawn from multiple distributions simultaneously. Jiang \textit{et al.} \cite{jiang2022learning} introduced a group distributionally robust optimization formulation for routing problems, aiming to learn policies that perform well across multiple predefined distribution groups. Bi \textit{et al.} \cite{bi2022learning} proposed AMDKD to distill knowledge from multiple teacher policies trained on various exemplar distributions, thereby improving robustness to unseen distributions. From the perspective of ensemble methods, Jiang \textit{et al.} \cite{jiang2023ensemble} learned a set of diverse sub-policies and ensemble them to mitigate performance drops under distribution shift, while Gao \textit{et al.} \cite{gao2024towards} proposed an ensemble-based framework with a transferrable local policy. However, a single model weight can hardly perform well across multiple tasks. Therefore, Manchanda \textit{et al.} \cite{manchanda2022generalization} and Zhou \textit{et al.} \cite{zhou2023towards} adopted meta-learning to learn a good initialization that can be efficiently fine-tuned afterward. Meanwhile, additional structural information, such as distance \cite{wang2025distance} and symmetricity \cite{kim2022sym}, has also been incorporated to enhance model generalization. Beyond cross-distribution generalization, some studies have also explored cross-scale \cite{fu2021generalize,hou2023generalize,son2023meta,luo2023neural,drakulic2023bq} and cross-problem generalization \cite{liu2024multi,zhou2024mvmoe}, and have achieved promising results.

\subsection{Mixture-of-Experts}

MoE models date back to modular learning with a gating function that routes inputs to specialized experts \cite{jacobs1991adaptive,jordan1994hierarchical}. Modern sparse MoE layers became influential in deep networks by enabling conditional computation, which only activate only a small subset of experts to increase effective capacity without proportional computation \cite{shazeer2017outrageously}. Subsequent work scaled MoE within Transformer training \cite{fedus2022switch} and focused on improving gating mechanism \cite{zhou2022mixture,puigcerver2023sparse,xue2024openmoe}, load balancing \cite{lewis2021base,nie2021evomoe} or applications in various domains \cite{carion2020end,you2022speechmoe2,mustafa2022multimodal}. As to neural methods for VRPs, Zhou \textit{et al.} made first attempt to combine MoE with multi-task learning to improve cross-problem generalization, which achieved promising results. For a more detailed review of MoE, we refer readers to the surveys \cite{yuksel2012twenty,mu2025comprehensive}.

\subsection{Motivation}

As discussed above, existing studies on cross-distribution generalization yield a fixed policy with a predetermined network architecture, regardless of the characteristics of the instance distribution. According to the No Free Lunch theorems \cite{wolpert2002no}, a single fixed policy is unrealistic to perform well across arbitrary distributions. To mitigate this issue, Zhou \textit{et al.} \cite{zhou2023towards} adopted meta-learning to learn a favorable initialization that can be easily fine-tuned on a target distribution. However, meta-learning typically trades optimality for fast adaptation, which still requires additional training on the target distribution at deployment.

Motivated by recent advances in LLMs, we explore MoE as a way to address cross-distribution generalization for VRPs. Specifically, we proposed R2E-IG, which activates the most suitable modules based on the distribution characteristics of each input instance, thereby enabling flexible policies across distributions. The effectiveness of MoE in solving VRPs has also been shown by Zhou \textit{et al.} \cite{zhou2024mvmoe}, who integrated MoE into multi-task learning to improve cross-problem generalization. In contrast to cross-problem generalization, where tasks are associated with explicit attributes, distribution characteristics are not directly observable and are harder to represent. To this end, we designed an instance-level gating mechanism that learns distribution-aware instance representations and activates appropriate modules accordingly. We further introduced a R2E architecture to enhance expert expressiveness via residual refinement. In addition, we proposed a mixed-distribution training strategy equipped with DWA to adaptively reweight training data across multiple distributions for more efficient learning. Extensive experiments have demonstrated the effectiveness of R2E-IG.

\begin{figure*}[!t]
	\centering
	\includegraphics[width=\linewidth]{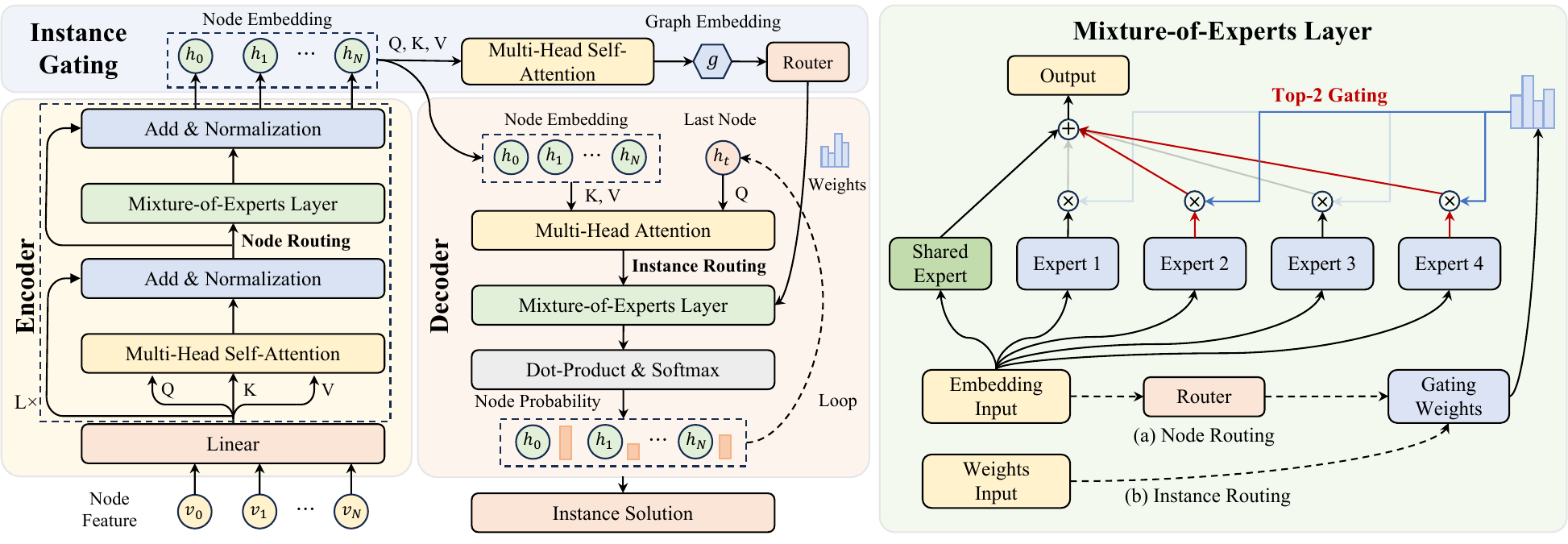}
	\caption{The model architecture of R2E-IG. [\textit{Yellow Part}]: Given a VRP instance, the encoder maps node features to node embeddings. [\textit{Blue Part}]: A graph embedding is aggregated from node embeddings via self-attention and fed into a lightweight router to produce instance-level routing weights. [\textit{Orange Part}]: The decoder constructs the solution auto-regressively by computing node-selection probabilities conditioned on the last selected node and the current partial solution. [\textit{Green Part}]: All FFN blocks in the encoder and decoder are replaced with MoE layers, supporting two routing types: node routing (gating experts by the current embedding) and instance routing (gating experts by the instance-level weights). We provide an example with Top-2 gating over 4 experts and a shared expert that is always activated.}
	\label{fig:R2E-IG}
\end{figure*}

\section{Preliminaries}
\label{PS}

\noindent In this section, we first present the VRP formulation and define the notion of distribution. We then describe how DRL-based methods solve VRPs.


\subsection{Problem Formulation}
We define a VRP instance of size $n$ as a graph $G=\{\mathcal{V},\mathcal{E}\}$, where the node set $\mathcal{V}$ consists of a depot node $v_0$ and $n$ customer nodes $\{v_i\}_{i=1}^n$, and the edge set $\mathcal{E}$ contains edges $e(v_i,v_j)$ between any pair of nodes $v_i$ and $v_j$. Each customer $v_i$ is associated with a demand $\delta_i$, and the vehicle has capacity $Q$. The solution (i.e., tour) $\tau$ is represented as an ordered sequence of nodes over $\mathcal{V}$, which can be decomposed into multiple sub-tours. We define the edge cost $C[e(v_i,v_j)]$ as the Euclidean distance between nodes $v_i$ and $v_j$. Accordingly, the tour length $S(\tau)$ is the sum of edge costs along $\tau$. The objective of VRP is to find an optimal tour $\tau^*$ with minimum length, as formulated in Eq. \ref{eq:VRP}.

\begin{equation}
\label{eq:VRP}
\tau^*=\underset{\tau^{\prime} \in \mathcal{S}}{\arg \min } S\left(\tau^{\prime} \mid \mathcal{G}\right)=\underset{\tau^{\prime} \in \mathcal{S}}{\arg \min } \sum_{e \in \tau^{\prime}} C\left[e\left(v_i, v_j\right)\right].
\end{equation}
where $\mathcal{S}$ denotes the set of all feasible solutions that satisfy the problem-specific constraints. In each sub-tour, the vehicle departs from the depot $v_0$, visits a subset of customers, and returns to the depot. A tour $\tau$ is feasible if each customer is visited exactly once and the total demand served on each sub-tour does not exceed the vehicle capacity $Q$.

\subsection{Distribution of Vehicle Routing Problems}

In this paper, we aim to enhance the model’s cross-distribution generalization ability, where distribution specifically refers to the spatial distribution of node coordinates in geographic space. Most prior studies generate training instances from a uniform distribution, whereas we train on multiple distributions (Uniform, Cluster, and Mixed) as ID data. We further evaluate on several unseen distributions as OoD test cases (Explosion, Expansion, Grid, and Implosion), following prior work \cite{bossek2019evolving,bi2022learning}. In addition, we test on benchmark datasets (e.g., CVRPLIB \cite{uchoa2017new} and TSPLIB \cite{reinelt1991tsplib}) as another form of unseen instances, which consider the real-world applications. Visualizations of these instance distributions are shown in Fig. \ref{fig:distribution}.

\subsection{Learn-for-Construction Methods}

Neural constructive methods typically formulate a VRP instance $\mathcal{G}$ as a Markov Decision Process (MDP) and build a solution sequentially in an auto-regressive manner. Representative studies such as AM \cite{kool2018attention} and POMO \cite{kwon2020pomo} adopt the encoder-decoder structure based on attention mechanism. The encoder maps the input nodes (e.g., coordinates and demands) into contextual node embeddings, capturing global instance structure via self-attention. Conditioned on these embeddings and the current partial route state (e.g., current node, remaining capacity, and visited set), the decoder outputs a probability distribution over feasible next nodes and selects the next node to append to the route, repeating this process until a complete solution is formed. During decoding, feasibility is enforced through masking, where the nodes violate constraints are masked so their probabilities are set to zero. The model can be viewed as a policy parameterized by $\theta$, and the probability of a solution is factorized by the chain rule as:

\begin{equation}
p_\theta(\tau \mid \mathcal{G})=\prod_{t=1}^T p_\theta\left(\pi_\theta(t) \mid \pi_\theta(<t), \mathcal{G}\right).
\end{equation}
where $\pi_\theta(t)$ and $\pi_\theta(<t)$ denote the selected node and current partial solution at the time step $t$, assuming a total of $T$ time steps. The reward is defined as the negative value of tour length, i.e., $\mathcal{R}=-L(\tau|\mathcal{G})$. To obtain a well-performed policy, the REINFORCE algorithm \cite{williams1992simple} is commonly adopted, which applies estimated gradients of the expected reward to optimize the network:

\begin{equation}
\small
\label{eq:alpha}
\nabla_\theta \mathcal{L}_{\alpha}(\theta \mid \mathcal{G})=\mathbb{E}_{p_\theta(\tau \mid \mathcal{G})}\left[(S(\tau)-b(\mathcal{G})) \nabla_\theta \log p_\theta(\tau \mid \mathcal{G})\right].
\end{equation}
where $b(\cdot)$ denotes the baseline, which can reduce gradient variance during training.

\section{Methodology}
\label{Methodology}

\noindent In this section, we present the details of R2E-IG, which consists of three key components: an MoE module with our proposed R2E architecture, an instance-level gating mechanism driven by self-learned instance representations, and a mixed-distribution training mechanism equipped with DWA. Our R2E-IG is built upon POMO \cite{kwon2020pomo} as the backbone and can be easily integrated into other DRL-based methods. The overall architecture of R2E-IG is shown in Fig. \ref{fig:R2E-IG}.

\subsection{Mixture-of-Experts Module}

A typical MoE layer consists of a lightweight router $G$ and $m$ expert modules $\{E_1,E_2,\ldots,E_m\}$. Given an input $x$, the router computes gating scores and performs sparse Top-$k$ routing, so that only $k$ experts with highest values are activated. The conditional computation preserves model capacity while reducing computation, and encourages various experts to specialize in capturing distribution-specific patterns. In addition, we introduce a shared expert $E_{sh}$ that is always activated to learn distribution-invariant knowledge across distributions. The MoE output is computed as a weighted combination of the activated experts:

\begin{equation}
\mathrm{MoE}(x) = E_{sh}(x) + \sum_{j\in\mathcal{K}} \tilde{G}(x)_j \cdot E_j(x).
\end{equation}
where $\mathcal{K}$ denotes the set of Top-$k$ activated experts, and $\tilde{G}(x)$ are the normalized gating weights over the activated experts.

\subsubsection{Lightweight Router}
Given an input representation $x$, the router first applies a linear projection to produce routing logits over $m$ experts. The logits are then converted into a probability vector $G(x)$ via the \texttt{Softmax} operator. To enforce sparse activation, we adopt Top-$k$ routing and only keep the $k$ experts with the largest values. Finally, we renormalize the retained probabilities over the selected experts to obtain the final gating weights, denoted as $\tilde{G}(x)$, which are used to weight the outputs of the activated experts.

\subsubsection{Residual Refined Expert}
\begin{figure}[t]
\centering
\begin{tabular}{cc}\hspace{-1.3cm}
\subfloat[Vanilla Expert.\label{fig:R2E:vanilla}]{
  \makebox[0.4\linewidth][c]{\includegraphics[height=3.2cm]{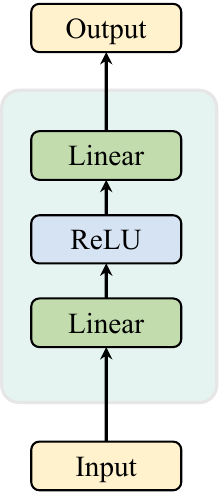}}
} \hspace{.1cm}
\subfloat[Residual Refined Expert.\label{fig:R2E:R2E}]{
  \makebox[0.4\linewidth][c]{\includegraphics[height=3.2cm]{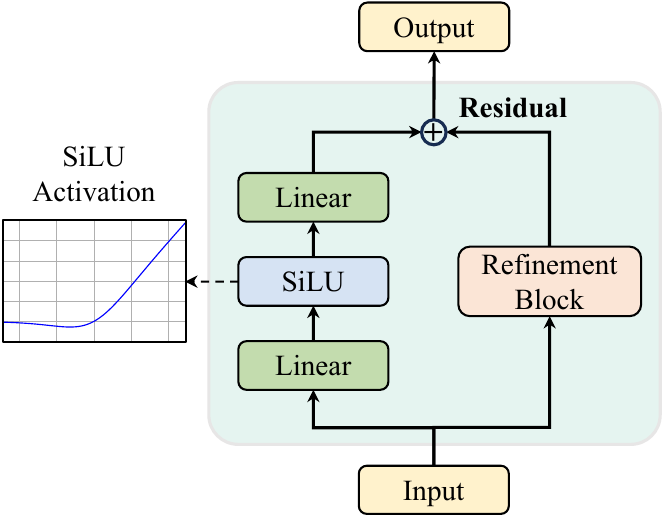}}
}
\end{tabular}
\caption{The architecture design of vanilla expert and R2E.}
\label{fig:R2E}
\end{figure}

In standard MoE implementations, each expert is typically instantiated as a two-layer feed-forward network. As illustrated in Fig. \ref{fig:R2E:vanilla}, a vanilla expert first projects the input features to a higher-dimensional space, applies a \texttt{ReLU} nonlinearity to increase expressiveness, and then projects back to the original feature space. As shown in Fig. \ref{fig:R2E:R2E}, for our R2E we replace \texttt{ReLU} with \texttt{SiLU} to obtain smoother activations and more stable optimization. More importantly, we introduce a residual refinement branch to strengthen the expert’s capacity with minimal extra cost. Specifically, in addition to the main path, we add a bypass that passes through a lightweight refinement block, implemented as a linear transformation in the original feature space. The refined residual is then added to the main expert output, enabling each expert to combine nonlinear transformation with feature-space refinement, which benefits generalization under distribution shift.

\subsection{Gating Mechanism}

In each MoE layer, the additional cost mainly comes from computing gating weights and dispatching nodes to the selected experts. While introducing MoE into the decoder can improve solution quality, decoding is performed step by step, and the number of decoding steps $T$ increases with the problem size $N$. In contrast, the encoder performs routing over a fixed number of nodes $N$ ($\ll T$), so that applying typical node-level routing in the decoder can incur substantially higher computation overhead. This motivates our instance-level gating mechanism, which better leverages MoEs in the decoder and achieves a favorable trade-off between empirical performance and computational cost. Next, we describe the node-level and instance-level gating mechanisms, respectively.

\subsubsection{Node-level Gating Mechanism}
Under node routing, we consider a single instance with node embeddings $X \in \mathbb{R}^{I \times d}$, where $I$ is the number of nodes that participate in the current forward pass and $d$ is the embedding dimension. The router computes a gating vector for each node embedding and activates the corresponding experts accordingly. This pipeline is straightforward and widely used, but it becomes computationally expensive when applied repeatedly during decoding. Therefore, we employ node routing only in the encoder.

\subsubsection{Instance-level Gating Mechanism}
To reduce the computational overhead, we apply instance routing in the decoder. Instead of computing gating weights for every node at each decoding step, instance routing derives the gating weights \emph{once} from a distribution-aware instance representation $z_{\mathrm{inst}}$, so that all nodes within the same instance share a unified expert activation pattern. Accordingly, the MoE takes node embeddings $X \in \mathbb{R}^{I \times d}$ together with predefined gating weights $\tilde{G}(z_{\mathrm{inst}})$ as inputs.

A straightforward choice for $z_{\mathrm{inst}}$ is to average node embeddings. However, our backbone \cite{kwon2020pomo} applies instance normalization over node embeddings, which largely removes the mean information across nodes and makes simple mean pooling less informative for characterizing the underlying instance distribution. To address this issue, we introduce an additional Multi-Head Attention (MHA) module over the node embeddings $X_{node}$ to learn a more expressive graph embedding:
\begin{equation}
\label{eq:inst_rep}
z_{\mathrm{inst}}=\mathrm{Pool}\big(\mathrm{MHA}(X_{node})\big).
\end{equation}
where $\mathrm{MHA}(\cdot)$ performs self-attention over nodes, and $\mathrm{Pool}(\cdot)$ is the mean pooling operator to obtain the instance-level representation. Specifically, we treat the construction of $z_{\mathrm{inst}}$ as an auxiliary distribution classification task to encourage it to capture distribution-relevant characteristics. We feed $z_{\mathrm{inst}}$ into a predictor to obtain predicted probability over $C$ predefined distribution categories among training:
\begin{equation}
\label{eq:predcition}
\hat{y}=\mathrm{Pred}(z_{\mathrm{inst}})\in\mathbb{R}^{C}.
\end{equation}
where the predictor is implemented as a single \texttt{Linear} layer followed by a \texttt{Softmax} operator. We then optimize the cross-entropy loss:
\begin{equation}
\label{eq:beta}
\mathcal{L}_{\beta}=-\sum_{c=1}^{C}y_c \log(\hat{y}_c).
\end{equation}
where $y$ is the distribution label. 
This auxiliary task encourages $z_{\mathrm{inst}}$ to become distribution-aware, and the resulting instance representation is then used to compute instance-level gating weights $\tilde{G}(z_{\mathrm{inst}})$, which are reused throughout decoding to avoid repeated per-node routing computations.

\subsubsection{Load Balancing Loss}
Sparse routing may lead to expert imbalance, where only a few experts are frequently selected while others are rarely utilized. To prevent such routing collapse and encourage balanced expert usage, we additionally adopt a load balancing auxiliary loss. Given the input node embeddings $X\in \mathbb{R}^{I\times d}$, the output probability of router can be descirbed as $G(X) \in \mathbb{R}^{I\times m}$, and we define the \emph{importance} of the $j$-th expert as the average probability for all nodes:
\begin{equation}
\label{eq:importance}
p_j = \frac{1}{I}\sum_{i=1}^{I} G(X)_{i,j},\quad \forall j=1,...,m.
\end{equation}
where $I$ is the total number of routed nodes. With Top-$k$ routing, we further define the \emph{load} of the $j$-th expert as the normalized frequency that it is actually selected:
\begin{equation}
\label{eq:load}
f_j = \frac{1}{Ik}\sum_{i=1}^{I}\mathbb{I}\!\left(j \in \mathcal{K}_i\right), \quad \forall j=1,...,m.
\end{equation}
where $\mathcal{K}_i$ denotes the set of Top-$k$ activated experts for the $i$-th node, and $\mathbb{I}\!\left(j \in \mathcal{K}_i\right)$ equals $1$ if the $j$-th expert is in the Top-$k$ set $\mathcal{K}_i$ for the $i$-th node, and $0$ otherwise. Since the importance $p$ and load $f$ are $m$-dimensional vectors, we can define the load balancing loss:

\begin{equation}
\label{eq:gamma}
\mathcal{L}_\gamma = m \sum_{j=1}^{m} p_j \cdot f_j.
\end{equation}
which penalizes mismatch between routing probabilities and actual expert utilization, encouraging all experts to be utilized more evenly. The load balancing loss are minimized to stabilize MoE training and improve expert diversity.
 
\subsection{Mixed-Distribution Training Mechanism}
As mentioned before, introducing multiple training distributions has become a common way for improving cross-distribution generalization in recent studies. Accordingly, how to manage and schedule data from various distributions during training is crucial. Bi \textit{et al.} \cite{bi2022learning} proposed an adaptive mechanism that switches the training distribution across epochs, allowing the model to focus on more informative distributions. However, frequent distribution switching may induce conflicting gradient directions and destabilize training. To address this issue, we propose a Mixed-Distribution Training Mechanism that jointly trains on multiple ID domains and employs DWA to adaptively adjust the mixture proportions across distributions.

\subsubsection{Distribution Setting}
We consider an ID set of distributions $\mathcal{D}=\{U,C,M\}$ (i.e., Uniform, Cluster, and Mixed). During training, each batch is constructed by sampling instances from these distributions according to an adaptive sampling probability $p_d$ for each $d\in\mathcal{D}$. To track performance change across ID domains, we construct distribution-specific validation sets $\mathcal{Z}_U$, $\mathcal{Z}_C$, and $\mathcal{Z}_M$, each containing 1,000 instances. At the end of each epoch, we evaluate the model on validation sets with the Gap metric. The Gap metric is defined as the relative difference between the tour length produced by the model and efficient solver, i.e., the ratio of their difference to the solver's tour length, which is always non-negative.

\subsubsection{Dynamic Weight Adaption}
As training progresses, the relative utility of various training distributions may change. Intuitively, a larger Gap indicates poorer performance on a distribution and thus calls for more training on that distribution. Similarly, a larger loss suggests higher uncertainty and potential room for improvement, implying that more instances from this distribution should be emphasized. Based on these intuitions, we define the sampling probability of each distribution to be proportional to the sum of its Gap and loss:

\begin{equation}
\small
\label{eq:weight}
p_d = \mathrm{Softmax}(\mathrm{AvgGap}[\tau_\theta,\tau_{solver}|\mathcal{Z}_d]+\mathcal{L}^d),\quad \forall d\in\mathcal{D}.
\end{equation}
where $\tau_\theta$ and $\tau_{solver}$ denote the solutions generated by the model $\theta$ and the LKH3 solver \cite{helsgaun2017extension}, respectively. $\mathrm{AvgGap}[\cdot]$ denotes the average Gap on the validation dataset, and $\mathcal{L}$ is the loss computed on training instances sampled from distribution $d$.

\subsubsection{Loss Function}
The overall loss consists of three components: (i) the task loss $\mathcal{L}_\alpha$ for solution construction trained by REINFORCE, (ii) the load balancing loss $\mathcal{L}_\beta$ for stabilizing sparse MoE routing, and (iii) an auxiliary  classification loss $\mathcal{L}_\gamma$ for learning distribution-aware instance representations. The total loss is formulated as:
\begin{equation}
\label{eq:total_loss} 
\mathcal{L} = \sum_{d\in \mathcal{D}} \mathcal{L}_\alpha^d + \omega_\beta\mathcal{L}_\beta^d + \omega_\gamma\mathcal{L}_\gamma^d.
\end{equation}
where $\omega_\beta$ and $\omega_\gamma$ control the relative weights of the corresponding loss terms, and $\mathcal{L}_\alpha$, $\mathcal{L}_\beta$, and $\mathcal{L}_\gamma$ are computed according to Eq.~\ref{eq:alpha}, Eq.~\ref{eq:beta}, and Eq.~\ref{eq:gamma}, respectively.

\begin{table*}[!t]
\caption{Cross-Distribution Generalization on In-Distribution Synthetic Instances.}
\label{tab:ID_results} 
\resizebox{\textwidth}{!}{
\begin{threeparttable}
\begin{tabular}{cc|ccccccccc|ccccccccc}
\toprule
\multirow{2}{*}{}        & \multirow{2}{*}{Method}        & \multicolumn{3}{c}{TSP20}                  & \multicolumn{3}{c}{TSP50}                   & \multicolumn{3}{c|}{TSP100}                & \multicolumn{3}{c}{CVRP20}                  & \multicolumn{3}{c}{CVRP50}                   & \multicolumn{3}{c}{CVRP100}                  \\
                         &                                & Obj.            & Gap$^*$               & Time$^*$ & Obj.            & Gap$^*$               & Time$^*$  & Obj.            & Gap$^*$               & Time$^*$ & Obj.            & Gap$^*$                & Time$^*$ & Obj.             & Gap$^*$               & Time$^*$  & Obj.             & Gap$^*$               & Time$^*$  \\ \midrule
\multirow{9}{*}{\rotatebox{90}{Uniform}} & \multicolumn{1}{c|}{LKH-3}     & 3.8448          & -                 & 1.3m & 5.6867          & -          & 14.3m & 7.7532          & -                 & 1.1h & 6.1637          & -                  & 3.0h & 10.4586          & -                 & 12.0h & 15.6751          & -                 & 22.6h \\
                         & \multicolumn{1}{c|}{POMO}      & \textbf{3.8448} & \textbf{0.0002\%} & 0.4s & 5.6881          & 0.0242\%          & 1.3s  & 7.7683          & 0.1940\%          & 5.1s & 6.1784          & 0.2382\%           & 0.5s & 10.5332          & 0.7131\%          & 1.7s  & 15.8129          & 0.8793\%          & 6.2s  \\
                         & \multicolumn{1}{c|}{DAR$^\#$}       & 3.8500          & 0.1357\%          & 0.3s & 5.6980          & 0.1990\%          & 1.1s  & 7.7936          & 0.5200\%          & 6.0s & 6.3087          & 2.3522\%           & 0.3s & 10.6686          & 2.0080\%          & 1.4s  & 16.0045          & 2.1017\%          & 7.4s  \\
                         & \multicolumn{1}{c|}{Omni-VRP$^\#$}  & 3.8589          & 0.3675\%          & 0.2s & 5.7363          & 0.8713\%          & 0.9s  & 7.8519          & 1.2729\%          & 4.9s & 6.5712          & 6.6111\%           & 0.3s & 10.8031          & 3.2938\%          & 1.2s  & 16.0222          & 2.2146\%          & 5.9s  \\
                         & \multicolumn{1}{c|}{ELG$^\#$}       & 3.8492          & 0.1138\%          & 0.4s & 5.6915          & 0.0840\%          & 2.5s  & 7.7717          & 0.2375\%          & 6.5s & 6.3940          & 3.7371\%           & 0.6s & 10.6389          & 1.7233\%          & 3.9s  & 15.8716          & 1.2538\%          & 9.5s  \\
                         & \multicolumn{1}{c|}{Sym-NCO}   & 3.8449          & 0.0013\%          & 0.7s & 5.6884          & 0.0292\%          & 2.0s  & 7.7728          & 0.2519\%          & 6.0s & 6.1790          & 0.2488\%           & 1.8s & 10.5848          & 1.2062\%          & 2.9s  & 15.8744          & 1.2712\%          & 7.7s  \\
                         & \multicolumn{1}{c|}{AMDKD}     & 3.8449          & 0.0028\%          & 0.4s & 5.6891          & 0.0420\%          & 1.3s  & 7.7837          & 0.3930\%          & 5.3s & 6.1875          & 0.3857\%           & 0.6s & 10.5403          & 0.7809\%          & 1.7s  & 15.8349          & 1.0197\%          & 6.3s  \\ \cmidrule(lr){2-20}
                         & \multicolumn{1}{c|}{R2E-IG}      & 3.8448          & 0.0005\%          & 0.6s & \textbf{5.6881} & \textbf{0.0242\%} & 2.5s  & \textbf{7.7665} & \textbf{0.1706\%} & 9.6s & \textbf{6.1719} & \textbf{0.1323\%}  & 1.1s & \textbf{10.5251} & \textbf{0.6358\%} & 3.6s  & \textbf{15.8082} & \textbf{0.8491\%} & 13.1s \\
                         & \multicolumn{1}{c|}{R2E-IG-SGBS$^\dagger$} & \textbf{3.8448} & \textbf{0.0000\%} & 2.4s & \textbf{5.6870} & \textbf{0.0040\%} & 17.2s & \textbf{7.7578} & \textbf{0.0588\%} & 1.5m & \textbf{6.1617} & \textbf{-0.0323\%} & 4.3s & \textbf{10.4742} & \textbf{0.1490\%} & 24.6s & \textbf{15.6932} & \textbf{0.1156\%} & 2.0m  \\ \midrule
\multirow{9}{*}{\rotatebox{90}{Cluster}} & \multicolumn{1}{c|}{LKH-3}     & 1.8249          & -                 & 1.8m & 2.6665          & -                 & 16.8m & 3.6678          & -                 & 1.6h & 3.0588          & -                  & 3.5h & 5.1908           & -                 & 14.2h & 7.8519           & -                 & 26.6h \\
                         & \multicolumn{1}{c|}{POMO}      & 1.8250          & 0.0050\%          & 0.4s & 2.6794          & 0.4831\%          & 1.3s  & 3.7472          & 2.1646\%          & 5.1s & 3.0667          & 0.2604\%           & 0.5s & 5.2628           & 1.3873\%          & 1.6s  & 8.0354           & 2.3363\%          & 7.1s  \\
                         & \multicolumn{1}{c|}{DAR$^\#$}       & 1.8319          & 0.3806\%          & 0.3s & 2.7058          & 1.4709\%          & 1.1s  & 3.7629          & 2.5925\%          & 5.9s & 3.1605          & 3.3258\%           & 0.3s & 5.3951           & 3.9358\%          & 1.5s  & 8.1990           & 4.4196\%          & 7.8s  \\
                         & \multicolumn{1}{c|}{Omni-VRP$^\#$}  & 1.8317          & 0.3713\%          & 0.2s & 2.6897          & 0.8707\%          & 1.0s  & 3.7188          & 1.3899\%          & 4.8s & 3.1683          & 3.5793\%           & 0.3s & 5.3261           & 2.6059\%          & 1.2s  & 8.0172           & 2.1046\%          & 5.9s  \\
                         & \multicolumn{1}{c|}{ELG$^\#$}       & 1.8310          & 0.3328\%          & 0.4s & 2.6883          & 0.8162\%          & 2.5s  & 3.7207          & 1.4412\%          & 6.4s & 3.1726          & 3.7209\%           & 0.6s & 5.3285           & 2.6533\%          & 4.3s  & 8.0760           & 2.8539\%          & 11.7s \\
                         & \multicolumn{1}{c|}{Sym-NCO}   & 1.8251          & 0.0112\%          & 0.7s & 2.6792          & 0.4746\%          & 1.7s  & 3.7531          & 2.3247\%          & 5.8s & 3.0679          & 0.2987\%           & 1.8s & 5.2946           & 1.9999\%          & 3.0s  & 8.0468           & 2.4817\%          & 8.5s  \\
                         & \multicolumn{1}{c|}{AMDKD}     & 1.8249          & 0.0016\%          & 0.4s & 2.6679          & 0.0528\%          & 1.4s  & 3.6795          & 0.3196\%          & 5.3s & 3.0719          & 0.4293\%           & 0.6s & 5.2382           & 0.9133\%          & 1.6s  & 7.9536           & 1.2952\%          & 6.3s  \\ \cmidrule(lr){2-20}
                         & \multicolumn{1}{c|}{R2E-IG}      & \textbf{1.8249} & \textbf{0.0012\%} & 0.6s & \textbf{2.6678} & \textbf{0.0478\%} & 2.5s  & \textbf{3.6791} & \textbf{0.3075\%} & 9.6s & \textbf{3.0625} & \textbf{0.1220\%}  & 1.1s & \textbf{5.2284}  & \textbf{0.7250\%} & 3.5s  & \textbf{7.9346}  & \textbf{1.0524\%} & 13.0s \\
                         & \multicolumn{1}{c|}{R2E-IG-SGBS$^\dagger$} & \textbf{1.8249} & \textbf{0.0000\%} & 2.5s & \textbf{2.6667} & \textbf{0.0063\%} & 17.1s & \textbf{3.6714} & \textbf{0.0983\%} & 1.5m & \textbf{3.0572} & \textbf{-0.0516\%} & 4.2s & \textbf{5.1994}  & \textbf{0.1647\%} & 24.9s & \textbf{7.8738}  & \textbf{0.2782\%} & 2.0m  \\ \midrule
\multirow{9}{*}{\rotatebox{90}{Mixed}}   & \multicolumn{1}{c|}{LKH-3}     & 3.2786          & -                 & 2.1m & 4.9113          & -                 & 20.8m & 6.7294          & -                 & 1.1h & 5.4950          & -                  & 3.6h & 9.4704           & -                 & 14.2h & 14.2125          & -                 & 26.2h \\
                         & \multicolumn{1}{c|}{POMO}      & 3.2787          & 0.0046\%          & 0.4s & 4.9188          & 0.1508\%          & 1.3s  & 6.7996          & 1.0417\%          & 5.1s & 5.5096          & 0.2651\%           & 0.6s & 9.5623           & 0.9701\%          & 1.6s  & 14.3902          & 1.2504\%          & 6.6s  \\
                         & \multicolumn{1}{c|}{DAR$^\#$}       & 3.2873          & 0.2656\%          & 0.3s & 4.9508          & 0.8037\%          & 1.1s  & 6.8366          & 1.5928\%          & 5.9s & 5.6630          & 3.0563\%           & 0.3s & 9.7067           & 2.4946\%          & 1.4s  & 14.5548          & 2.4083\%          & 7.3s  \\
                         & \multicolumn{1}{c|}{Omni-VRP$^\#$}  & 3.3117          & 1.0105\%          & 0.2s & 5.0002          & 1.8094\%          & 1.1s  & 6.8781          & 2.2089\%          & 4.8s & 5.7903          & 5.3726\%           & 0.3s & 9.7367           & 2.8113\%          & 1.2s  & 14.5161          & 2.1363\%          & 5.9s  \\
                         & \multicolumn{1}{c|}{ELG$^\#$}       & 3.2879          & 0.2830\%          & 0.4s & 4.9326          & 0.4325\%          & 2.5s  & 6.7972          & 1.0069\%          & 6.5s & 5.7399          & 4.4567\%           & 0.6s & 9.6599           & 2.0005\%          & 3.9s  & 14.4556          & 1.7107\%          & 9.9s  \\
                         & \multicolumn{1}{c|}{Sym-NCO}   & 3.2788          & 0.0076\%          & 0.8s & 4.9189          & 0.1545\%          & 1.8s  & 6.8146          & 1.2650\%          & 5.9s & 5.5153          & 0.3688\%           & 1.8s & 9.6110           & 1.4840\%          & 3.0s  & 14.4480          & 1.6567\%          & 7.8s  \\
                         & \multicolumn{1}{c|}{AMDKD}     & \textbf{3.2787}          & \textbf{0.0025}\%          & 0.4s & 4.9151          & 0.0766\%          & 1.4s  & \textbf{6.7531} & \textbf{0.3509\%} & 5.4s & 5.5151          & 0.3655\%           & 0.5s & 9.5498           & 0.8385\%          & 1.7s  & 14.3524          & 0.9845\%          & 6.2s  \\ \cmidrule(lr){2-20}
                         & \multicolumn{1}{c|}{R2E-IG}      & 3.2787 & 0.0033\% & 0.6s & \textbf{4.9140} & \textbf{0.0537\%} & 2.5s  & 6.7540          & 0.3649\%          & 9.7s & \textbf{5.5029} & \textbf{0.1435\%}  & 1.1s & \textbf{9.5267}  & \textbf{0.5937\%} & 4.0s  & \textbf{14.3164} & \textbf{0.7312\%} & 12.8s \\
                         & \multicolumn{1}{c|}{R2E-IG-SGBS$^\dagger$} & \textbf{3.2786} & \textbf{0.0004\%} & 2.4s & \textbf{4.9122} & \textbf{0.0173\%} & 17.3s & \textbf{6.7417} & \textbf{0.1820\%} & 1.5m & \textbf{5.4941} & \textbf{-0.0177\%} & 4.1s & \textbf{9.4849}  & \textbf{0.1523\%} & 24.3s & \textbf{14.2259} & \textbf{0.0946\%} & 2.0m  \\ \bottomrule
\end{tabular}
\begin{tablenotes} 
\footnotesize
\item $^*$ We report the total time for solving the dataset, each containing 1,000 instances, and `-' represents 0.000\%, with which the gaps are computed.
\item $^\#$ The corresponding results are derived by open-sourced pretrained models, while other models are retrained under our experimental settings.
\item $^\dagger$ For SGBS, we set the beam width $\beta=4$ and expansion factor $\gamma=4$.
\end{tablenotes} 
\end{threeparttable} 
}
\end{table*}

\section{Experimental Results}
\label{exp}
\noindent In this section, we conduct comprehensive experiments to evaluate the effectiveness of R2E-IG. We first detail the experimental settings. Next, we compare R2E-IG with state-of-the-art baselines on synthetic datasets under both ID and OoD instances. We then further validate its cross-distribution generalization performance on benchmark datasets with real-world instance distributions. Finally, we perform ablation studies to assess the contribution of each proposed component and further analyze the impact of various hyper-parameter configurations.

\subsection{Experimental Settings}
We conduct experiments on Capacitated Vehicle Routing Problems (CVRP) and Traveling Salesman Problem (TSP) under the problems scale $n=20,50,100$. Following AMDKD \cite{bi2022learning}, we adopt Uniform, Cluster, and Mixed (mixture of Uniform and Cluster) as exemplar distributions during training, while Explosion, Expansion, Grid, and Implosion are used as unseen distributions as testing. For convenience and fairness, we directly test on the dataset\footnote{\url{https://drive.google.com/drive/folders/1-Jf1Rj88zPHWoUlj71ssRiX52b6Ex0Q9}} released by Bi \textit{et al.}, and report results on the first 1,000 instances. All experiments are conducted on a machine with NVIDIA RTX 4090 GPU cards and Intel(R) Xeon(R) Gold 6230 CPU @ 2.10GHz.

\subsubsection{Training}

\begin{figure}[!t]
	\centering
	\includegraphics[width=\columnwidth]{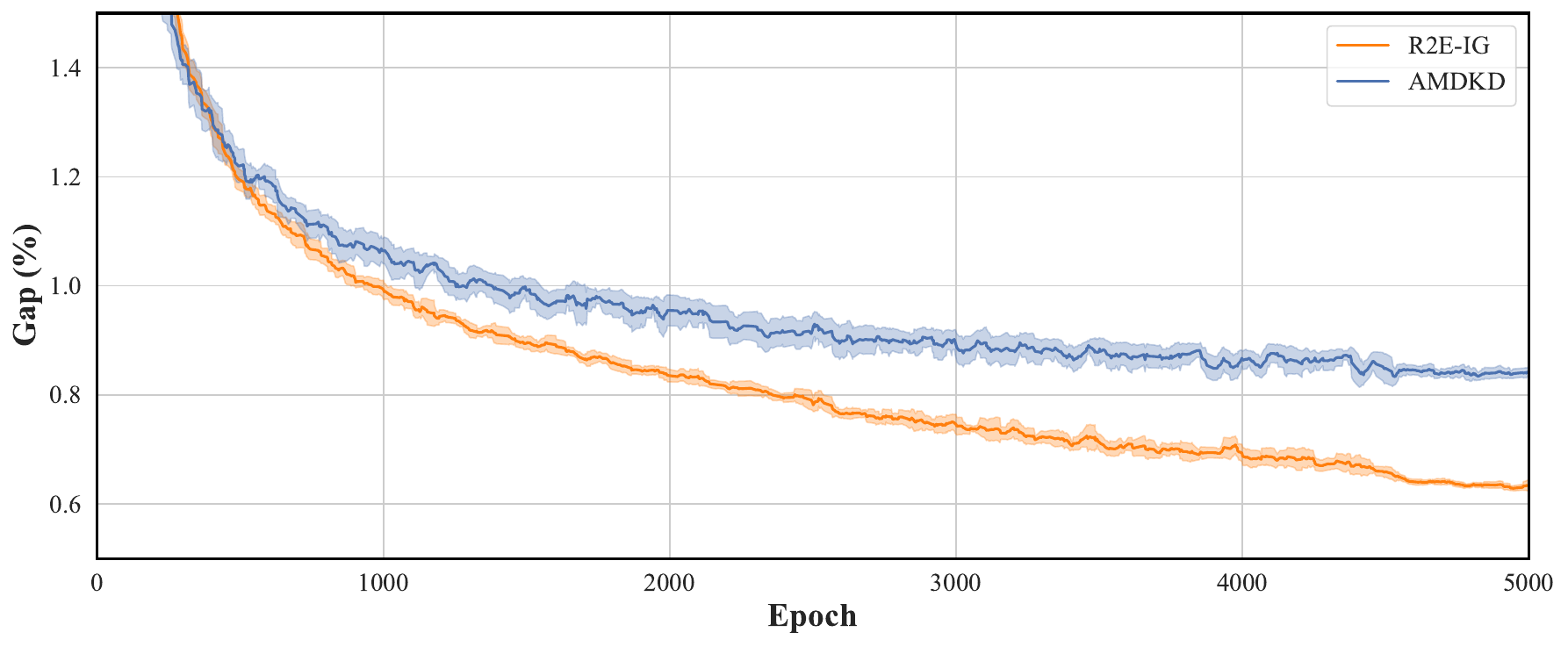}
	\caption{Training curves on CVRP50, where the average Gap is computed over all ID validation sets (i.e., $\mathcal{Z}_U$, $\mathcal{Z}_C$, and $\mathcal{Z}_M$). R2E-IG converges faster and exhibits lower variance than AMDKD.}
	\label{fig:training}
\end{figure}

We adopt POMO \cite{kwon2020pomo} as the backbone of R2E-IG, which is one of the most representative DRL-based methods for VRPs. We use the Adam optimizer with a learning rate of $10^{-4}$, weight decay of $10^{-6}$, and a batch size of 256 across all problem scales. The model is trained for 5,000 epochs, each containing 20,000 training instances. Specifically, the learning rate will decay by 10 for the last 10\% training instances (i.e., decay at epoch 4500). We set the embedding dimension to 128, use 8 attention heads, and adopt a 6-layer encoder. For the MoE configuration, we employ $m=8$ experts with Top-$k$ routing ($k=3$), and set $\omega_\beta=0.1$ and $\omega_\gamma=0.01$. For each expert, the intermediate dimension is set to 128. In addition, we provide the learning curves of the average Gap on all ID validation sets throughout training for a representative case of CVRP with $n=50$. Our method exhibits notably better training stability compared with AMDKD, as evidenced by a smaller variance of the validation Gap across epochs, as shown in Fig.~\ref{fig:training}. Moreover, R2E-IG converges faster and reaches a lower Gap at convergence, indicating more efficient optimization and improved final solution quality.

\subsubsection{Inference}
For all neural solvers, we adopt greedy decoding with $8\times$ instance augmentation following POMO \cite{kwon2020pomo}. We report the average objective value and gap on each dataset, along with the total runtime for solving the full set of instances. Specifically, the gaps are computed with respect to the results of the best-performing traditional VRP solvers (i.e., `-' in Tables \ref{tab:ID_results} and \ref{tab:OoD_results}).

\subsubsection{Baseline Algorithms}

To comprehensively evaluate the performance, we compare R2E-IG with state-of-the-art baseline algorithms. We take the LKH3 solver \cite{helsgaun2017extension} as the heuristic baseline, which is one of the most representative and solid solver for both TSP and CVRP. We also compare with the backbone algorithm POMO \cite{kwon2020pomo}, which is trained on only Uniform distribution with the same training settings. Meanwhile, other representative DRL-based methods specialized for generalization are considered, including: 1) DAR \cite{wang2025distance}, a novel model leverages distance information to adjust attention scores, thereby enhancing the generalization; 2) Omni-VRP \cite{zhou2023towards}, a meta-learning framework that learns a good initialization of model parameters for fine-tuning afterwards; 3) ELG \cite{gao2024towards}, an ensemble-based framework with transferrable local policy, where aggregated policies perform cooperatively and complementarily to boost generalization; 4) Sym-NCO \cite{kim2022sym}, a novel training scheme that utilizes symmetricities (i.e., rotational and reflectional invariance) to improve the generalization capability; 5) AMDKD \cite{bi2022learning}, an efficient framework leverages various knowledge from multiple exemplar distributions to yield a generalist model. Furthermore, our R2E-IG can be easily coupled with recent Simulated-Guided Beam Search (SGBS) \cite{choo2022simulation}, denoted as R2E-IG-SGBS, which can achieve superior performance without additional modifications.

\begin{table*}[!t]
\caption{Cross-Distribution Generalization on Out-of-Distribution Synthetic Instances.}
\label{tab:OoD_results} 
\resizebox{\textwidth}{!}{
\begin{threeparttable}
\begin{tabular}{cc|ccccccccc|ccccccccc}
\toprule
\multirow{2}{*}{}                              & \multirow{2}{*}{Method}        & \multicolumn{3}{c}{TSP20}                   & \multicolumn{3}{c}{TSP50}                   & \multicolumn{3}{c|}{TSP100}                 & \multicolumn{3}{c}{CVRP20}                  & \multicolumn{3}{c}{CVRP50}                   & \multicolumn{3}{c}{CVRP100}                  \\
                                               &                                & Obj.            & Gap$^*$               & Time$^*$  & Obj.            & Gap$^*$               & Time$^*$  & Obj.            & Gap$^*$               & Time$^*$  & Obj.            & Gap$^*$                & Time$^*$ & Obj.             & Gap$^*$               & Time$^*$  & Obj.             & Gap$^*$               & Time$^*$  \\ \midrule
\multirow{9}{*}{\rotatebox{90}{Expansion}}                     & \multicolumn{1}{c|}{LKH-3}     & 3.4361          & -                 & 2.8m  & 4.3892          & -                 & 26.1m & 5.3977          & -                 & 2.8h  & 5.3802          & -                  & 3.1h & 8.2143           & -                 & 12.2h & 11.3866          & -                 & 22.4h \\
                                               & \multicolumn{1}{c|}{POMO}      & \textbf{3.4361} & \textbf{0.0008\%} & 0.6s  & \textbf{4.3920} & \textbf{0.0643\%} & 1.8s  & 5.4369          & 0.7271\%          & 6.3s  & 5.3925          & 0.2278\%           & 0.7s & 8.2839           & 0.8479\%          & 2.2s  & 11.5387          & 1.3356\%          & 7.7s  \\
                                               & \multicolumn{1}{c|}{DAR}       & 3.4405          & 0.1299\%          & 0.3s  & 4.4124          & 0.5300\%          & 1.1s  & 5.4637          & 1.2237\%          & 5.9s  & 5.5006          & 2.2380\%           & 0.3s & 8.4037           & 2.3063\%          & 1.5s  & 11.6807          & 2.5829\%          & 7.5s  \\
                                               & \multicolumn{1}{c|}{Omni-VRP}  & 3.4528          & 0.4862\%          & 0.2s  & 4.4348          & 1.0385\%          & 1.1s  & 5.4791          & 1.5077\%          & 4.8s  & 5.6607          & 5.2134\%           & 0.3s & 8.4371           & 2.7133\%          & 1.2s  & 11.6234          & 2.0799\%          & 5.9s  \\
                                               & \multicolumn{1}{c|}{ELG}       & 3.4409          & 0.1404\%          & 0.4s  & 4.3979          & 0.1989\%          & 2.5s  & 5.4309          & 0.6150\%          & 6.5s  & 5.5624          & 3.3859\%           & 0.6s & 8.3589           & 1.7604\%          & 4.3s  & 11.5869          & 1.7591\%          & 11.2s \\
                                               & \multicolumn{1}{c|}{Sym-NCO}   & 3.4361          & 0.0026\%          & 0.9s  & 4.3927          & 0.0803\%          & 2.2s  & 5.4412          & 0.8065\%          & 6.6s  & 5.3939          & 0.2550\%           & 2.0s & 8.3231           & 1.3248\%          & 3.4s  & 11.5720          & 1.6281\%          & 8.6s  \\
                                               & \multicolumn{1}{c|}{AMDKD}     & 3.4361          & 0.0022\%          & 0.7s  & 4.3923          & 0.0707\%          & 1.9s  & 5.4294          & 0.5876\%          & 6.6s  & 5.4012          & 0.3892\%           & 0.7s & 8.2865           & 0.8800\%          & 2.1s  & 11.5416          & 1.3615\%          & 7.0s  \\ \cmidrule(lr){2-20}
                                               & \multicolumn{1}{c|}{R2E-IG}      & 3.4361          & 0.0018\%          & 0.8s  & 4.3920          & 0.0654\%          & 3.1s  & \textbf{5.4242} & \textbf{0.4909\%} & 10.7s & \textbf{5.3864} & \textbf{0.1145\%}  & 1.2s & \textbf{8.2697}  & \textbf{0.6750\%} & 4.2s  & \textbf{11.5078} & \textbf{1.0649\%} & 14.9s \\
                                               & \multicolumn{1}{c|}{R2E-IG-SGBS} & \textbf{3.4361} & \textbf{0.0000\%} & 2.7s  & \textbf{4.3897} & \textbf{0.0117\%} & 18.3s & \textbf{5.4063} & \textbf{0.1603\%} & 1.5m  & \textbf{5.3767} & \textbf{-0.0645\%} & 4.3s & \textbf{8.2284}  & \textbf{0.1723\%} & 25.1s & \textbf{11.4175} & \textbf{0.2710\%} & 2.0m  \\ \midrule
\multirow{9}{*}{\rotatebox{90}{Explosion}}                     & \multicolumn{1}{c|}{LKH-3}     & 3.5710          & -                 & 1.6m  & 4.6364          & -                 & 14.4m & 5.8404          & -                 & 1.1h  & 5.7918          & -                  & 3.1h & 8.7904           & -                 & 12.3h & 12.2973          & -                 & 22.2h \\
                                               & \multicolumn{1}{c|}{POMO}      & \textbf{3.5710} & \textbf{0.0000\%} & 0.6s  & 4.6377          & 0.0280\%          & 1.7s  & 5.8540          & 0.2327\%          & 6.4s  & 5.8022          & 0.1795\%           & 0.6s & 8.8542           & 0.7255\%          & 2.1s  & 12.4272          & 1.0565\%          & 7.3s  \\
                                               & \multicolumn{1}{c|}{DAR}       & 3.5746          & 0.1016\%          & 0.3s  & 4.6495          & 0.2822\%          & 1.1s  & 5.8826          & 0.7220\%          & 5.9s  & 5.9240          & 2.2837\%           & 0.3s & 8.9798           & 2.1551\%          & 1.4s  & 12.5887          & 2.3700\%          & 7.3s  \\
                                               & \multicolumn{1}{c|}{Omni-VRP}  & 3.5815          & 0.2931\%          & 0.2s  & 4.6677          & 0.6765\%          & 0.9s  & 5.9024          & 1.0620\%          & 4.8s  & 6.1395          & 6.0029\%           & 0.3s & 9.0515           & 2.9704\%          & 1.2s  & 12.5738          & 2.2488\%          & 5.9s  \\
                                               & \multicolumn{1}{c|}{ELG}       & 3.5740          & 0.0853\%          & 0.4s  & 4.6402          & 0.0823\%          & 2.5s  & 5.8535          & 0.2242\%          & 6.5s  & 5.9922          & 3.4595\%           & 0.6s & 8.9348           & 1.6427\%          & 4.0s  & 12.4762          & 1.4550\%          & 9.8s  \\
                                               & \multicolumn{1}{c|}{Sym-NCO}   & 3.5710          & 0.0005\%          & 0.9s  & 4.6378          & 0.0318\%          & 2.3s  & 5.8589          & 0.3177\%          & 6.9s  & 5.8047          & 0.2236\%           & 2.0s & 8.8950           & 1.1896\%          & 3.5s  & 12.4586          & 1.3118\%          & 8.6s  \\
                                               & \multicolumn{1}{c|}{AMDKD}     & 3.5710          & 0.0005\%          & 0.6s  & 4.6379          & 0.0325\%          & 1.8s  & 5.8556          & 0.2600\%          & 6.5s  & 5.8096          & 0.3070\%           & 0.7s & 8.8600           & 0.7921\%          & 2.2s  & 12.4481          & 1.2266\%          & 7.2s  \\ \cmidrule(lr){2-20}
                                               & \multicolumn{1}{c|}{R2E-IG}      & 3.5710          & 0.0003\%          & 0.9s  & \textbf{4.6376} & \textbf{0.0261\%} & 3.0s  & \textbf{5.8498} & \textbf{0.1605\%} & 10.7s & \textbf{5.7965} & \textbf{0.0813\%}  & 1.3s & \textbf{8.8399}  & \textbf{0.5638\%} & 4.1s  & \textbf{12.4102} & \textbf{0.9186\%} & 13.7s \\
                                               & \multicolumn{1}{c|}{R2E-IG-SGBS} & \textbf{3.5710} & \textbf{0.0000\%} & 2.6s  & \textbf{4.6366} & \textbf{0.0052\%} & 17.7s & \textbf{5.8427} & \textbf{0.0403\%} & 1.5m  & \textbf{5.7889} & \textbf{-0.0504\%} & 4.3s & \textbf{8.8002}  & \textbf{0.1114\%} & 25.1s & \textbf{12.3139} & \textbf{0.1357\%} & 2.0m  \\ \midrule
\multirow{9}{*}{\rotatebox{90}{Grid}}                          & \multicolumn{1}{c|}{LKH-3}     & 3.8796          & -                 & 1.6m  & 5.7013          & -                 & 15.3m & 7.7905          & -                 & 1.1h  & 6.1970          & -                  & 3.1h & 10.4515          & -                 & 12.1h & 15.6063          & -                 & 22.1h \\
                                               & \multicolumn{1}{c|}{POMO}      & 3.8796          & 0.0008\%          & 0.6s  & 5.7029          & 0.0293\%          & 1.9s  & 7.8060          & 0.1991\%          & 6.2s  & 6.2120          & 0.2422\%           & 0.7s  & 10.5269          & 0.7215\%          & 2.1s  & 15.7446          & 0.8859\%          & 7.4s  \\
                                               & \multicolumn{1}{c|}{DAR}       & 3.8838          & 0.1082\%          & 0.3s  & 5.7130          & 0.2058\%          & 1.1s  & 7.8310          & 0.5204\%          & 6.0s  & 6.3356          & 2.2372\%           & 0.3s & 10.6592          & 1.9875\%          & 1.4s  & 15.9280          & 2.0611\%          & 7.3s  \\
                                               & \multicolumn{1}{c|}{Omni-VRP}  & 3.8925          & 0.3326\%          & 0.2s  & 5.7516          & 0.8827\%          & 0.9s  & 7.8902          & 1.2806\%          & 4.8s  & 6.6162          & 6.7641\%           & 0.3s & 10.7936          & 3.2726\%          & 1.4s  & 15.9448          & 2.1689\%          & 5.9s  \\
                                               & \multicolumn{1}{c|}{ELG}       & 3.8835          & 0.1014\%          & 0.44s & 5.7058          & 0.0798\%          & 2.5s  & 7.8074          & 0.2179\%          & 6.5s  & 6.4286          & 3.7380\%           & 0.6s & 10.6287          & 1.6953\%          & 3.9s  & 15.7997          & 1.2393\%          & 9.5s  \\
                                               & \multicolumn{1}{c|}{Sym-NCO}   & 3.8797          & 0.0025\%          & 1.0s  & 5.7030          & 0.0300\%          & 2.2s  & 7.8118          & 0.2732\%          & 6.5s  & 6.2129          & 0.2561\%           & 1.9s & 10.5723          & 1.1557\%          & 3.5s  & 15.7985          & 1.2315\%          & 8.4s  \\
                                               & \multicolumn{1}{c|}{AMDKD}     & 3.8797          & 0.0022\%          & 0.6s  & 5.7039          & 0.0470\%          & 1.9s  & 7.8206          & 0.3865\%          & 6.4s  & 6.2208          & 0.3841\%           & 0.7s & 10.5355          & 0.8038\%          & 2.2s  & 15.7599          & 0.9843\%          & 7.1s  \\ \cmidrule(lr){2-20}
                                               & \multicolumn{1}{c|}{R2E-IG}      & \textbf{3.8796} & \textbf{0.0005\%} & 0.9s  & \textbf{5.7027} & \textbf{0.0247\%} & 3.0s  & \textbf{7.8050} & \textbf{0.1862\%} & 10.8s & \textbf{6.2049} & \textbf{0.1281\%}  & 1.2s & \textbf{10.5137} & \textbf{0.5953\%} & 4.0s  & \textbf{15.7409} & \textbf{0.8623\%} & 13.9s \\
                                               & \multicolumn{1}{c|}{R2E-IG-SGBS} & \textbf{3.8796} & \textbf{0.0000\%} & 2.7s  & \textbf{5.7015} & \textbf{0.0039\%} & 17.8s & \textbf{7.7955} & \textbf{0.0643\%} & 1.5m  & \textbf{6.1948} & \textbf{-0.0357\%} & 4.2s & \textbf{10.4649} & \textbf{0.1285\%} & 25.2s & \textbf{15.6268} & \textbf{0.1315\%} & 2.0m  \\ \midrule
\multirow{9}{*}{\rotatebox{90}{Implosion}} & \multicolumn{1}{c|}{LKH-3}     & 3.8207          & -                 & 1.5m  & 5.6123          & -                 & 15.1m & 7.6071          & -                 & 1.2h  & 6.0831          & -                  & 3.0h & 10.2900          & -                 & 12.1h & 15.4331          & -                 & 22.1h \\
\multicolumn{1}{l}{}                           & \multicolumn{1}{c|}{POMO}      & 3.8207          & 0.0012\%          & 0.6   & 5.6142          & 0.0336\%          & 1.8s  & 7.6223          & 0.1995\%          & 6.4s  & 6.0974          & 0.2355\%           & 0.7s & 10.3625          & 0.7046\%          & 2.0s  & 15.5674          & 0.8706\%          & 3.0s  \\
\multicolumn{1}{l}{}                           & \multicolumn{1}{c|}{DAR}       & 3.8247          & 0.1058\%          & 0.2s  & 5.6248          & 0.2224\%          & 1.1s  & 7.6483          & 0.5407\%          & 6.0s  & 6.2243          & 2.3226\%           & 0.3s & 10.4931          & 1.9742\%          & 1.4s  & 15.7461          & 2.0282\%          & 7.4s  \\
\multicolumn{1}{l}{}                           & \multicolumn{1}{c|}{Omni-VRP}  & 3.8331          & 0.3261\%          & 0.2s  & 5.6620          & 0.8850\%          & 0.9s  & 7.7009          & 1.2319\%          & 5.4s  & 6.5057          & 6.9477\%           & 0.3s & 10.6284          & 3.2891\%          & 1.4s  & 15.7656          & 2.1548\%          & 5.9s  \\
\multicolumn{1}{l}{}                           & \multicolumn{1}{c|}{ELG}       & 3.8245          & 0.0996\%          & 0.4s  & 5.6176          & 0.0941\%          & 2.5s  & 7.6243          & 0.2260\%          & 6.5s  & 6.3189          & 3.8762\%           & 0.6s & 10.4671          & 1.7215\%          & 3.9s  & 15.6182          & 1.1994\%          & 9.5s  \\
\multicolumn{1}{l}{}                           & \multicolumn{1}{c|}{Sym-NCO}   & 3.8207          & 0.0020\%          & 0.9s  & 5.6140          & 0.0310\%          & 2.1s  & 7.6267          & 0.2569\%          & 6.5s  & 6.0984          & 0.2517\%           & 2.0s & 10.4132          & 1.1974\%          & 3.5s  & 15.6222          & 1.2254\%          & 8.5s  \\
\multicolumn{1}{l}{}                           & \multicolumn{1}{c|}{AMDKD}     & 3.8208          & 0.0034\%          & 0.6s  & 5.6149          & 0.0464\%          & 1.9s  & 7.6368          & 0.3897\%          & 6.7s  & 6.1041          & 0.3457\%           & 0.7s & 10.3691          & 0.7690\%          & 2.2s  & 15.5833          & 0.9735\%          & 7.2s  \\ \cmidrule(lr){2-20}
\multicolumn{1}{l}{}                           & \multicolumn{1}{c|}{R2E-IG}      & \textbf{3.8207} & \textbf{0.0011\%} & 0.8s  & \textbf{5.6140} & \textbf{0.0307\%} & 3.0s  & \textbf{7.6202} & \textbf{0.1710\%} & 10.8s & \textbf{6.0907} & \textbf{0.1263\%}  & 1.3s & \textbf{10.3501} & \textbf{0.5840\%} & 4.2s  & \textbf{15.5601} & \textbf{0.8228\%} & 14.0s \\
\multicolumn{1}{l}{}                           & \multicolumn{1}{c|}{R2E-IG-SGBS} & \textbf{3.8207} & \textbf{0.0004\%} & 2.6s  & \textbf{5.6127} & \textbf{0.0078\%} & 17.8s & \textbf{7.6112} & \textbf{0.0536\%} & 1.5m  & \textbf{6.0801} & \textbf{-0.0480\%} & 4.2s & \textbf{10.3022} & \textbf{0.1184\%} & 25.1s & \textbf{15.4486} & \textbf{0.1002\%} & 2.0m  \\ \bottomrule
\end{tabular}
\begin{tablenotes} 
\footnotesize
\item $^*$ We report the total time for solving the dataset, each containing 1,000 instances, and `-' represents 0.000\%, with which the gaps are computed.
\item $^\#$ The corresponding results are derived by open-sourced pretrained models, while other models are retrained under our experimental settings.
\item $^\dagger$ For SGBS, we set the beam width $\beta=4$ and expansion factor $\gamma=4$.
\end{tablenotes} 
\end{threeparttable} 
}
\end{table*}

\subsection{Comparison Study on Synthetic Datasets}
\subsubsection{In-Distribution Performance}

Table~\ref{tab:ID_results} reports the results on ID synthetic instances for both TSP and CVRP across three training distributions (Uniform, Cluster, and Mixed). For the easiest setting TSP20, the task is relatively simple and most baselines already achieve near-optimal performances, where the differences are minor. Although POMO and AMDKD outperform R2E-IG on TSP20-Uniform and TSP20-Mixed, the advantage is marginal. As the problem scale increases (i.e., $n=50,100$), the benefits of R2E-IG becomes more evident, yielding consistently better results with improved stability. Across all three ID distributions, R2E-IG remains competitive against DRL-based baselines, which aligns with our design goal. By jointly training on multiple ID distributions, R2E-IG is encouraged to recognize distribution-specific characteristics and to specialize its computation through sparse expert activation. Specifically, in Section~\ref{sec:pattern}, we further investigate how expert activation patterns vary across different instance distributions.

Notably, POMO is trained solely on Uniform instances and is thus expected to be more specialized for this distribution. Nevertheless, R2E-IG still outperforms POMO in most settings even on Uniform test instances, while other baselines are all inferior to POMO on the Uniform distribution. The result indicate that our approach not only improves cross-distribution generalization but also enhances performance on a single distribution. Meanwhile, integrating SGBS further improves solution quality at the cost of slightly increased runtime, and even surpasses LKH3 in some cases (e.g., CVRP20), demonstrating a favorable trade-off between accuracy and efficiency.

\subsubsection{Out-of-Distribution Performance}

We also evaluate cross-distribution generalization on four unseen OoD synthetic distributions (Expansion, Explosion, Grid, and Implosion), as summarized in Table~\ref{tab:OoD_results}. Consistent with ID results, performance differences on TSP20 are generally marginal due to the relatively low difficulty. As the problem scale increases to $n=50$ and $n=100$, the performance differences among methods become more evident, and R2E-IG exhibits clearer advantages across most settings. Overall, R2E-IG achieves the best or highly competitive performance among DRL-based baselines on the majority of OoD cases, indicating strong robustness under distribution shift. We attribute this improvement to the MoE-based modularization and instance-level gating, which activates distribution-relevant experts together with a shared expert for common patterns. Specifically, R2E-IG can adapt its effective computation to the underlying spatial structure of each instance, rather than relying on a single fixed set of modules. Consequently, the distribution knowledge learned from ID training data transfers effectively to OoD instances, demonstrating strong cross-distribution generalization even on previously unseen distributions.

\begin{table*}[!t]
\caption{Performance on Benchmark Instances with Real-World Distribution.}
\label{tab:bench_results} 
\resizebox{\textwidth}{!}{
\begin{threeparttable}

\begin{tabular}{cc|cccccc|cc|cc|cccccc|cc}
\toprule
\multicolumn{2}{c|}{TSPLIB}   & \multirow{2}{*}{ELG} & \multirow{2}{*}{DAR} & \multirow{2}{*}{Omni-VRP} & \multirow{2}{*}{Sym-NCO} & \multirow{2}{*}{POMO} & \multirow{2}{*}{AMDKD} & \multirow{2}{*}{R2E-IG} & \multirow{2}{*}{R2E-IG-SGBS} & \multicolumn{2}{c|}{CVRPLIB}  & \multirow{2}{*}{ELG} & \multirow{2}{*}{DAR} & \multirow{2}{*}{Omni-VRP} & \multirow{2}{*}{Sym-NCO} & \multirow{2}{*}{POMO} & \multirow{2}{*}{AMDKD} & \multirow{2}{*}{R2E-IG} & \multirow{2}{*}{R2E-IG-SGBS} \\
Instance       & Opt.$^*$         &                      &                      &                           &                          &                       &                        &                       &                            & Instance         & Opt.$^*$       &                      &                      &                           &                          &                       &                        &                       &                            \\ \midrule
KroA100        & 21282        & 21346                & 21990                & 21305                     & 21436                    & 21644                 & \textbf{21285}         & \textbf{21285}        & \textbf{21285}             & X-n101-k25       & 27591      & 28971                & \textbf{28167}       & 29442                     & 29242                    & 29013                 & 28731                  & 29165                 & 28390                      \\
KroB100        & 22141        & 22283                & 22955                & 22650                     & 22502                    & 22319                 & \textbf{22197}         & 22263                 & \textbf{22197}             & X-n106-k14       & 26362      & 27344                & 27213                & 26990                     & 27138                    & 27900                 & 26812                  & \textbf{26684}        & \textbf{26682}             \\
KroC100        & 20749        & 20770                & 21187                & 20902                     & 20899                    & 20944                 & \textbf{20752}         & 20755                 & \textbf{20751}             & X-n110-k13       & 14971      & 15252                & 15731                & 15285                     & 15349                    & \textbf{15045}        & 15134                  & 15115                 & \textbf{15045}             \\
KroD100        & 21294        & 21531                & 21928                & 21828                     & 21780                    & 21751                 & \textbf{21307}         & 21388                 & \textbf{21294}             & X-n115-k10       & 12747      & 13307                & 13342                & 13240                     & 13407                    & 13509                 & \textbf{13189}         & 13419                 & \textbf{12978}             \\
KroE100        & 22068        & 22238                & 22671                & 22239                     & \textbf{22161}           & 22448                 & 22171                  & 22177                 & \textbf{22157}             & X-n120-k6        & 13332      & 13812                & \textbf{13663}       & 13944                     & 13877                    & 13820                 & 13722                  & 13821                 & \textbf{13632}             \\
rd100          & 7910         & 7917                 & 8045                 & 7958                      & 7949                     & \textbf{7916}         & 7921                   & 7944                  & 7944                       & X-n125-k30       & 55539      & 58303                & \textbf{57310}       & 58738                     & 58408                    & 59608                 & 58244                  & 58187                 & 57511                      \\
lin105         & 14379        & 14463                & 14893                & 14819                     & 14600                    & 14687                 & 14428                  & \textbf{14406}        & \textbf{14383}             & X-n129-k18       & 28940      & 29530                & 29361                & 29975                     & 29473                    & 29571                 & \textbf{29286}         & 29486                 & \textbf{29222}             \\
pr124          & 59030        & 59181                & 59713                & 59238                     & 59133                    & 59202                 & 59133                  & \textbf{59075}        & \textbf{59031}             & X-n134-k13       & 10916      & 11348                & 11484                & 11302                     & 11390                    & 11248                 & 11225                  & \textbf{11123}        & \textbf{11117}             \\
bier127        & 118282       & 122844               & 125586               & 121129                    & 124243                   & 124069                & 119111                 & \textbf{118949}       & \textbf{118770}            & X-n139-k10       & 13590      & 13873                & 14040                & 14019                     & 14139                    & \textbf{13846}        & 14107                  & 14027                 & \textbf{13820}             \\
ch130          & 6110         & 6117                 & 6229                 & 6251                      & 6153                     & \textbf{6115}         & 6150                   & 6131                  & 6125                       & X-n143-k7        & 15700      & \textbf{15993}       & 16310                & 16602                     & 16302                    & 16061                 & 16098                  & 16314                 & 16019                      \\
pr136          & 96772        & 97730                & 98945                & 97780                     & 98254                    & \textbf{97710}        & 98438                  & 97764                 & \textbf{96994}             & X-n148-k46       & 43448      & 45536                & \textbf{44705}       & 46438                     & 46075                    & 46307                 & 45887                  & 45705                 & 45014                      \\
pr144          & 58537        & 58860                & 59239                & 59571                     & 59328                    & 58724                 & 59343                  & \textbf{58617}        & \textbf{58535}             & X-n157-k13       & 16876      & 17634                & 17648                & \textbf{17107}            & 18380                    & 17459                 & 17841                  & 17321                 & 17121                      \\
ch150          & 6528         & 6581                 & 6602                 & 6586                      & 6635                     & 6571                  & 6624                   & \textbf{6566}         & \textbf{6559}              & X-n162-k11       & 14138      & 14614                & 14829                & \textbf{14595}            & 15134                    & 14660                 & 14848                  & 14996                 & 14675                      \\
kroA150        & 26524        & 26806                & 27617                & 26873                     & 27104                    & 26942                 & \textbf{26612}         & 26667                 & \textbf{26580}             & X-n167-k10       & 20557      & \textbf{21195}       & 21452                & 21436                     & 21617                    & 21324                 & 21431                  & 21340                 & \textbf{21082}             \\
kroB150        & 26130        & 26394                & 27167                & 26452                     & 26482                    & 26655                 & 26362                  & \textbf{26222}        & \textbf{26141}             & X-n172-k51       & 45607      & 48476                & \textbf{46682}       & 48399                     & 48311                    & 49616                 & 48104                  & 48645                 & 47029                      \\
pr152          & 73682        & \textbf{73718}       & 73995                & 74907                     & 75242                    & 74953                 & 74821                  & 74032                 & 73841                      & X-n176-k26       & 47812      & 52210                & \textbf{50074}       & 51332                     & 51126                    & 54833                 & 52679                  & 52045                 & 51662                      \\
u159           & 42080        & \textbf{42477}       & 42542                & 42561                     & 42525                    & 42546                 & 42498                  & 42498                 & \textbf{42470}             & X-n186-k15       & 24145      & 24808                & 25486                & \textbf{24768}            & 25798                    & 25541                 & 25663                  & 25436                 & 25055                      \\
d198           & 15780        & 17202                & 17898                & 16280                     & 17952                    & 18893                 & 16630                  & \textbf{16222}        & \textbf{16055}             & X-n190-k8        & 16980      & 17996                & 17814                & 17645                     & 18196                    & 18102                 & 17399                  & \textbf{17345}        & \textbf{17209}             \\
kroA200        & 29368        & 29835                & 30397                & 29823                     & 30148                    & 29910                 & \textbf{29816}         & 29845                 & \textbf{29494}             & X-n195-k51       & 44225      & 46825                & \textbf{45760}       & 47477                     & 47326                    & 48243                 & 47414                  & 47738                 & 46175                      \\
kroB200        & 29437        & 29902                & 30718                & 29814                     & 30427                    & 30297                 & 29800                  & \textbf{29623}        & \textbf{29469}             & X-n200-k36       & 58578      & 61535                & \textbf{60829}       & 61496                     & 61824                    & 62116                 & 61207                  & 61133                 & 60687                      \\ \midrule
\multicolumn{2}{c|}{Avg. Gap} & 1.27\%               & 3.24\%               & 1.51\%                    & 2.24\%                   & 2.38\%                & 0.96\%                 & \textbf{0.61\%}       & \textbf{0.31\%}            & \multicolumn{2}{c|}{Avg. Gap} & 4.08\%               & 4.20\%               & 4.04\%                    & 5.36\%                   & 5.55\%                & 3.93\%                 & \textbf{3.80\%}       & \textbf{2.38\%}            \\ \bottomrule
\end{tabular}
\begin{tablenotes} 
\footnotesize
\item $^*$ `Opt.' denotes the optimal value of the instance.
\end{tablenotes} 
\end{threeparttable} 
}
\end{table*}

\subsection{Comparison Study on Real-World Benchmarks}

Beyond the synthetic datasets, we further evaluate the cross-distribution generalization on real-world benchmarks (e.g., CVRPLIB \cite{uchoa2017new} and TSPLIB \cite{reinelt1991tsplib}), whose instance distributions deviate substantially from the synthetic ones. Specifically, for each problem, we select 20 representative benchmark instances with sizes ranging from 100 to 200 nodes. Specifically, most baselines (i.e., ELG, DAR, and Omni-VRP) incorporate additional design choices or training mechanisms to handle cross-distribution generalization, whereas we directly evaluate the model trained on $n=100$ without any distribution-specific adaptation. As shown in Table~\ref{tab:bench_results}, R2E-IG achieves the best average performance among all baselines on both benchmarks, reducing the average Gap to 0.61\% on TSPLIB and 3.80\% on CVRPLIB. The results indicate that the distribution-aware characteristics learned from mixed synthetic training distributions can transfer effectively to realistic instances. Although R2E-IG is not the best-performing method on some instances, it achieves the lowest average Gap across all benchmark instances. Therefore, R2E-IG is more stable and robust, rather than excelling only on a subset of instances. R2E-IG delivers consistently strong performance over diverse instances and distributions, making it a reliable method in realistic settings where instance characteristics can vary substantially. In addition, integrating SGBS consistently yields further improvements, bringing the average Gap down to 0.31\% on TSPLIB and 2.38\% on CVRPLIB.

\subsection{Gating Analysis}
The previous results have shown that R2E-IG outperforms strong baselines, while it remains important to evaluate whether the model can truly recognize distribution-aware characteristics and accordingly activate specialized experts.

\begin{figure}[t]
\centering
\setlength{\tabcolsep}{1pt}
\vspace{-.6cm}
\begin{tabular}{cc}
\hspace{-.4cm}
\subfloat[CVRP50]{\includegraphics[width=.48\columnwidth]{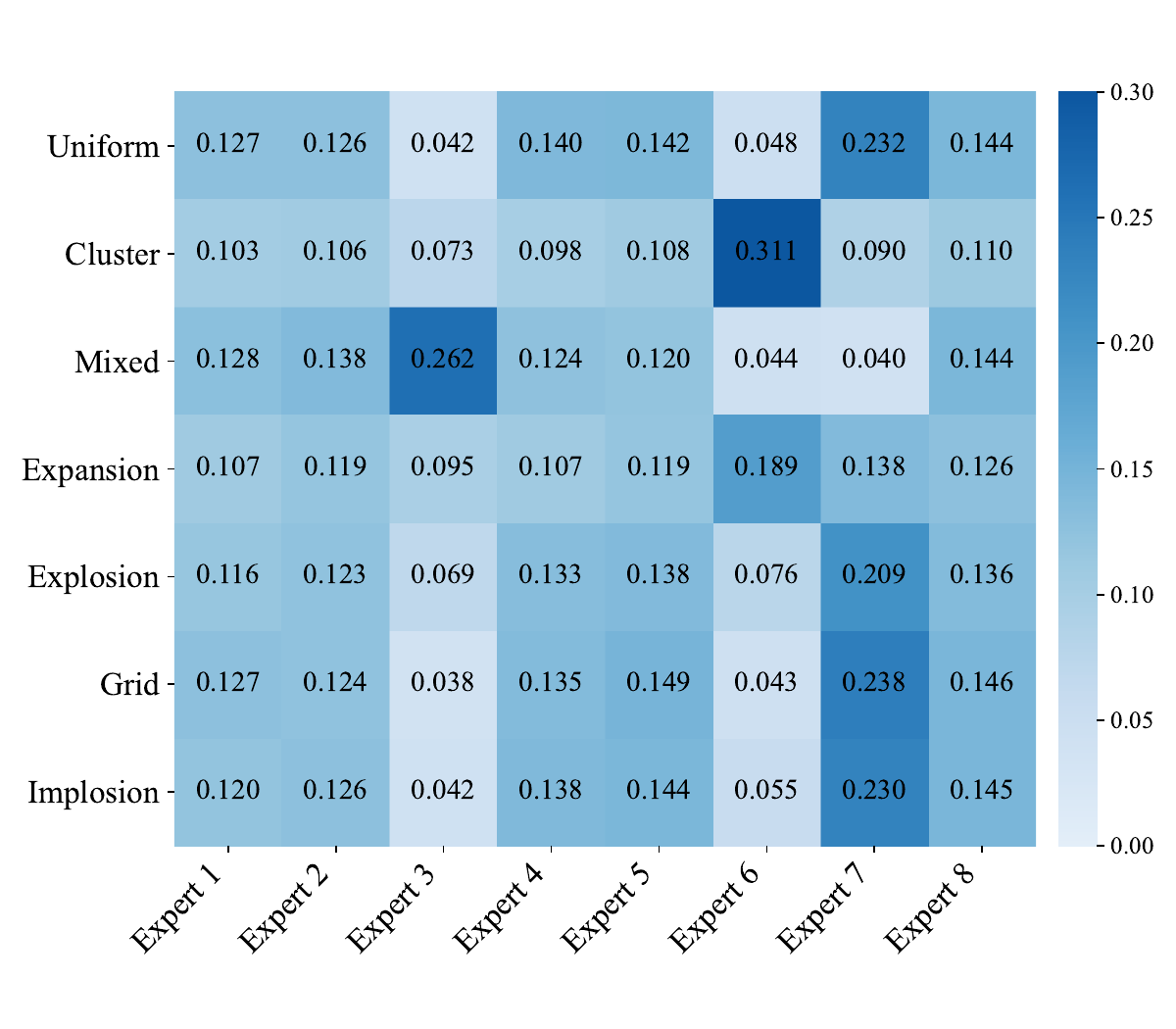}} &
\subfloat[CVRP100]{\includegraphics[width=.48\columnwidth]{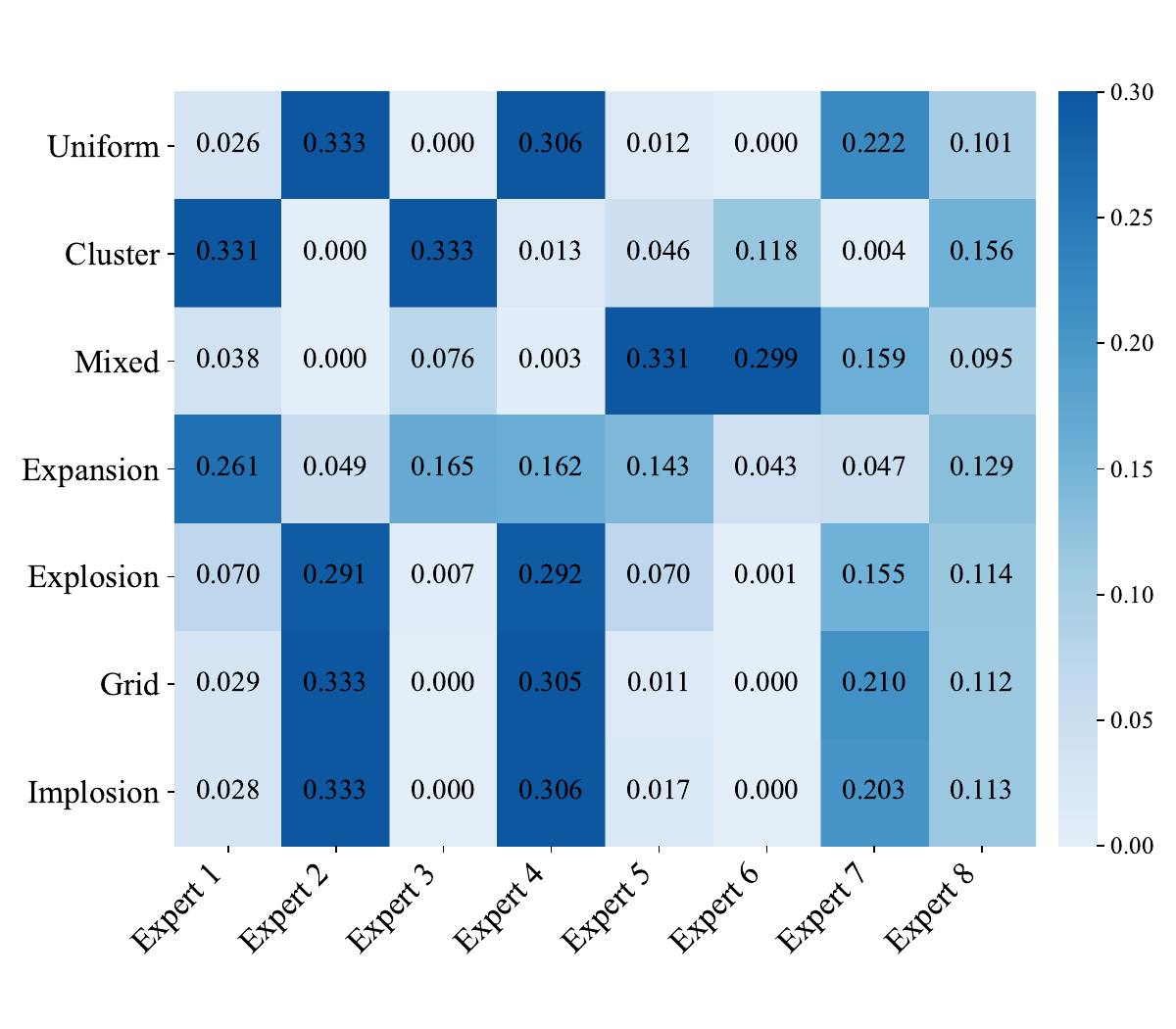}}
\end{tabular}
\caption{The expert activation patterns corresponding to various distributions.}
\label{fig:pattern}
\end{figure}

\subsubsection{Expert Activation Pattern Visualization}
\label{sec:pattern}

The previous results show that R2E-IG outperforms strong baselines, while it remains important to evaluate whether the model can truly recognize distribution-aware characteristics and accordingly activate specialized experts. Since distributional discrepancies typically become more noticeable as the problem scale increases, we select CVRP50 and CVRP100 as representative test cases. Under our setting, Top-$k$ routing activates $k=3$ experts from $m=8$ candidates for each instance. Intuitively, if the model effectively leverages distribution information, we expect different distributions to induce different expert combinations, rather than consistently selecting the same experts. Therefore, for each distribution, we count how often each expert is selected across 1,000 instances and visualize the resulting frequencies as heatmaps, as shown in Fig.~\ref{fig:pattern}.

As the problem scale increases from 50 to 100, the underlying distributional differences become more pronounced, which leads to more distinctive expert activation patterns (i.e., higher contrast in the heatmaps). For the three ID distributions (Uniform, Cluster, and Mixed), the high-frequency expert sets differ substantially. For example, on CVRP100, Uniform tends to favor experts 2/4/7, whereas Cluster more often activates experts 1/3/8, indicating that the model learns distribution-dependent gating behaviors. On OoD instances, the activation patterns between Expansion and Cluster are relatively close, indicating that both share common distribution-aware characteristics (i.e., clustered spatial structures). In contrast, Explosion, Grid, and Implosion show expert combinations more similar to Uniform. A plausible explanation is that, at $n=100$, the structural deviations of these distributions from Uniform are not sufficiently pronounced to yield sharply separated instance representations, leading to similar routing outcomes.

\begin{figure}[!t]
	\centering
	\includegraphics[width=\columnwidth]{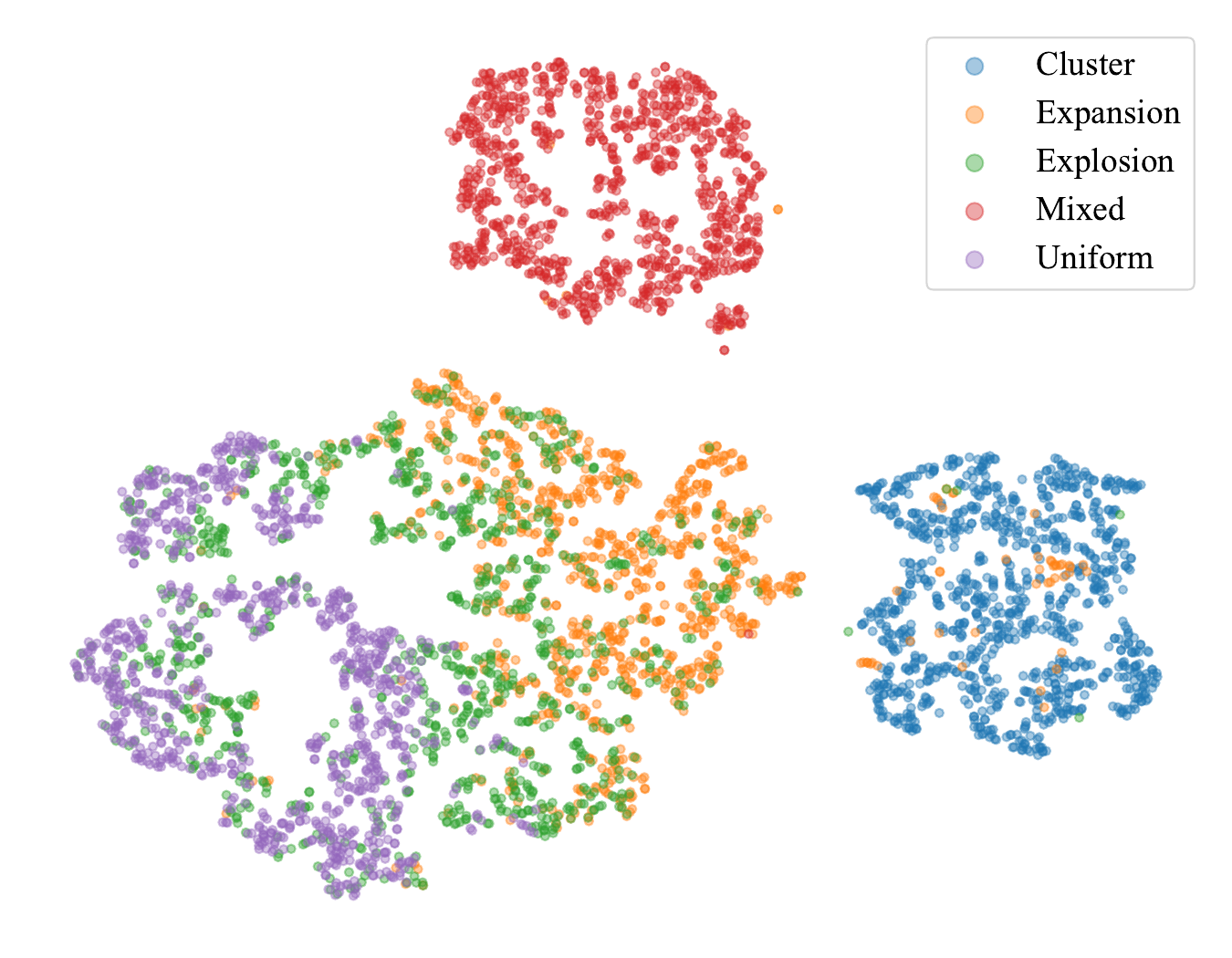}
	\vspace{-.5cm}
	\caption{Visualization of distribution-specific graph representations using t-SNE, where each distribution comprises 1,000 instances.}
	\label{fig:tsne}
\end{figure}

\begin{figure*}[!t]
\centering
\setlength{\tabcolsep}{1pt}
\vspace{-.6cm}
\begin{tabular}{ccccc}
\subfloat[Expert Design\label{fig:ablation:expert}]{\includegraphics[width=.2\textwidth]{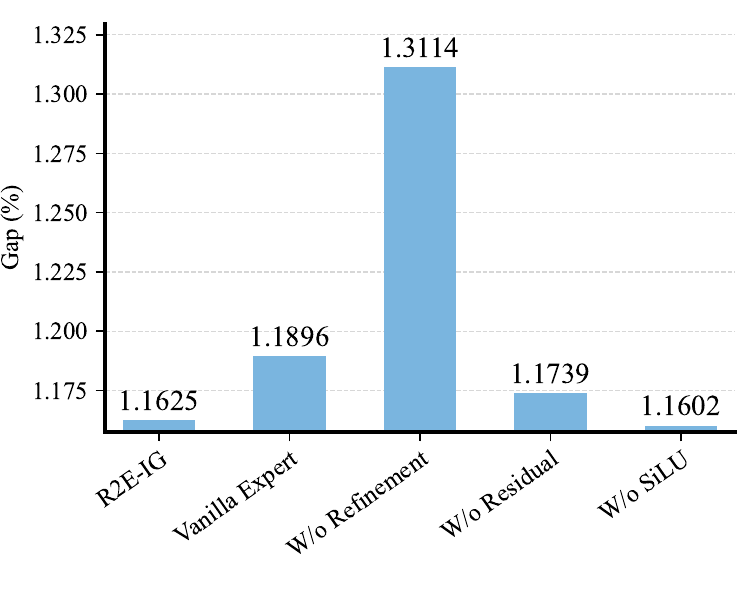}} &
\subfloat[Training Configuration\label{fig:ablation:training}]{\includegraphics[width=.2\textwidth]{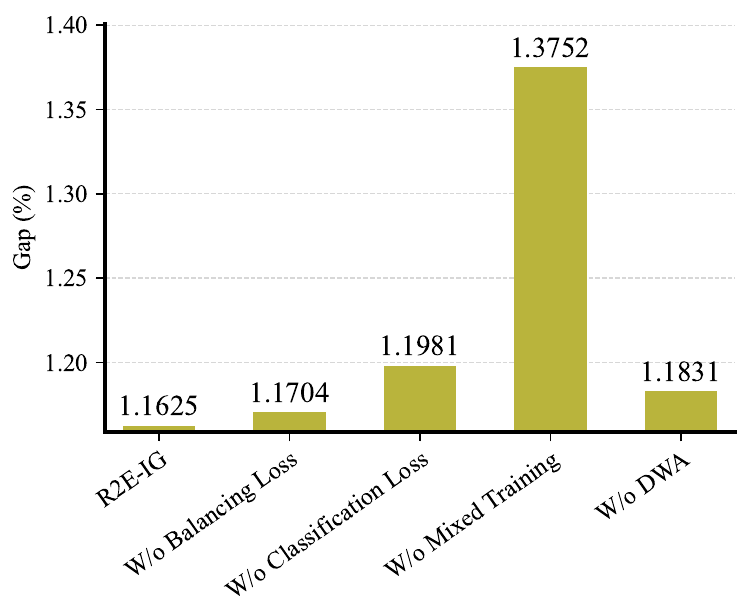}} & 
\subfloat[MoE Configurations]{\includegraphics[width=.2\textwidth]{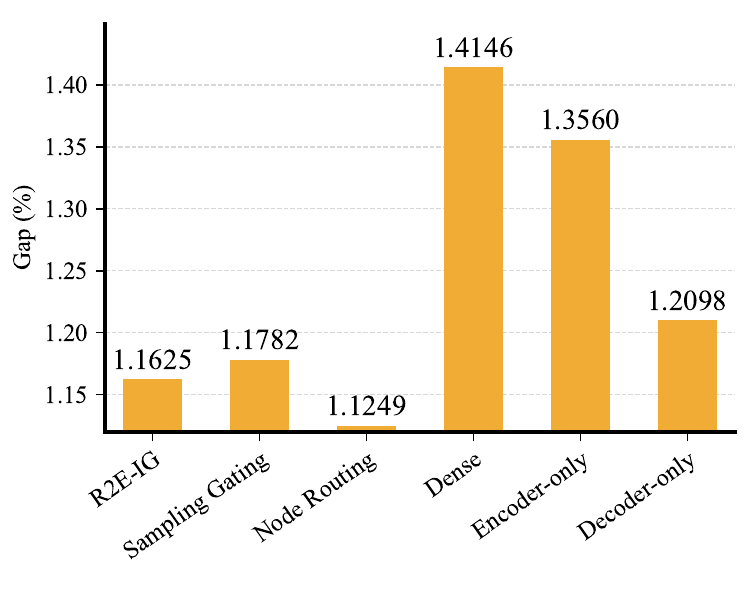}} & 
\subfloat[Number of Experts]{\includegraphics[width=.2\textwidth]{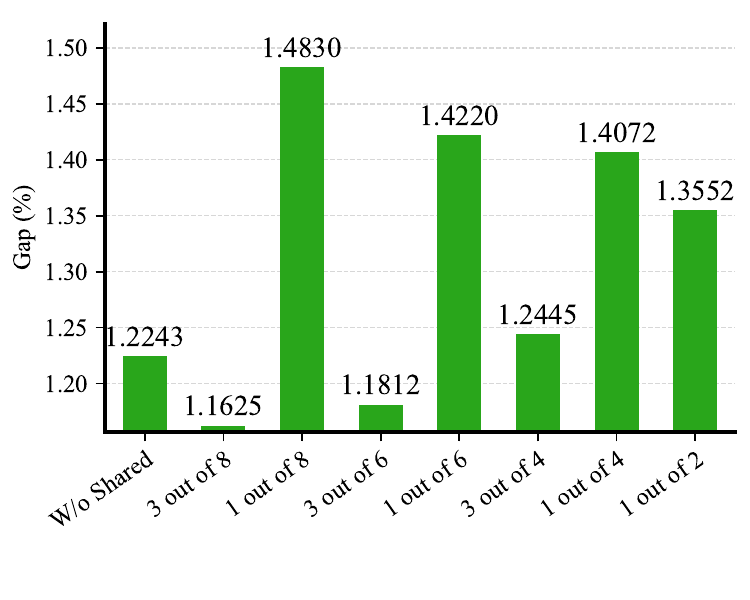}} & 
\subfloat[Dimension of Experts]{\includegraphics[width=.2\textwidth]{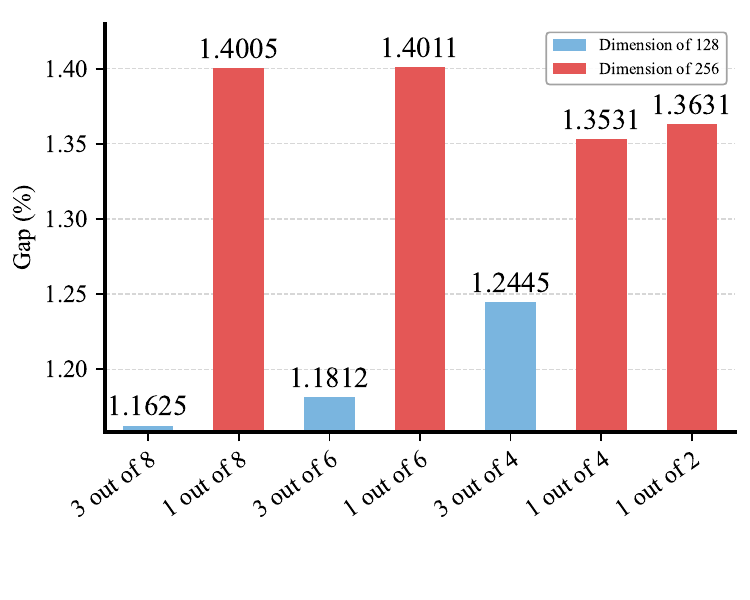}} 
\end{tabular}
\caption{The effect of different configurations on average performance, where the Gap is averaged over all seven ID and OoD distribution datasets. The first three panels (a)--(c) present ablations of the proposed components, while the last two panels (d)–-(e) examine the impact of expert configurations in terms of the number of experts and expert dimension.}
\label{fig:ablation}
\end{figure*}

\subsubsection{t-SNE Visualization}

We visualize the learned representations with t-SNE \cite{maaten2008visualizing}, which projects high-dimensional features into a two-dimensional space. Specifically, we apply t-SNE to the graph embeddings of all ID instances (Uniform, Cluster, and Mixed) as well as representative OoD instances (Expansion and Explosion), using 1,000 instances for each distribution. As shown in Fig. \ref{fig:tsne}, the results demonstrate clear boundaries between distributions, indicating that the model learns distributional-specific representations. We also observe partial overlap between some distributions, reflecting shared spatial characteristics. For example, Uniform, Explosion, and Expansion exhibit intersections, consistent with their broadly dispersed node layouts. Moreover, a subset of Expansion instances is encompassed by Cluster instances, which intuitive because the Expansion process intensifies spatial concentration. As the concentration becomes stronger, Expansion can degenerate into a clustered structure, leading to similar instance-level representations.

\subsection{Ablation Study and Further Analysis}

We conduct an ablation study to assess the contribution of each proposed component and to examine how various MoE configurations affect the performance of R2E-IG. To save the computational cost, we reduce the total number of training epochs from 5,000 to 1,000, each with 10,0000 training instances. Specifically, the Gap is averaged over all seven ID and OoD distribution datasets, and the results are shown in Fig. \ref{fig:ablation}.

\subsubsection{Expert Design}

We propose a novel expert structure R2E, and first compare it with the vanilla expert design. We then conduct ablations by removing the refinement block (\emph{W/o Refinement}) or the entire residual bypass (\emph{W/o Residual}). In addition, we replace \texttt{SiLU} with \texttt{ReLU} (\emph{W/o SiLU}) to further assess the contribution of the activation choice. The results highlight that the refinement block is crucial. Simply introducing a residual bypass without the refinement block can even degrade performance compared with a vanilla expert, suggesting that an unrefined shortcut may inject noisy or misaligned feature updates. In addition, while \texttt{SiLU} and \texttt{ReLU} yield very similar average Gaps when aggregated over all seven ID and OoD datasets, \texttt{SiLU} achieves a lower average Gap on OoD instances. This indicates that \texttt{SiLU} provides better representational robustness, thereby enhancing generalization to unseen distributions. Most importantly, our R2E outperforms the vanilla expert, reducing the average Gap from 1.1896\% to 1.1625\%.

\subsubsection{Training Configuration}

For training techniques, we incorporate two auxiliary objectives $\mathcal{L_\beta}$ and $\mathcal{L_\gamma}$), and further propose a mixed-distribution training mechanism equipped with DWA. Accordingly, we isolate each component and conduct ablation studies to quantify its individual contribution. The results show that each individual component contributes positively to the performance of R2E-IG, with the proposed mixed-distribution training mechanism providing the most substantial gains.

\subsubsection{MoE Configuration}

We investigate MoE configurations from three aspects: the gating scheme (\emph{Top-$k$ Gating} or \emph{Sampling Gating}), the routing mode of decoder (\emph{Instance Routing} or \emph{Node Routing}), and the positions of applying MoE modules (\emph{Dense}, \emph{Encoder-only}, \emph{Decoder-only} or \emph{Encoder-Decoder}). Compared to \emph{Top-$k$ Gating} adopted in R2E-IG, \emph{Sampling Gating} samples $k$ experts to activate based on the gating weights, which may include extra variance and yield unstable expert patterns. \emph{Node Routing} computes expert assignments at the node level, making it more flexible than \emph{Instance Routing}. However, the improved performance comes with a substantially higher computational cost because routing must be performed repeatedly for a large number of nodes during decoding. Meanwhile, we observe that, compared with a dense static architecture, introducing MoE layers in either the encoder or the decoder consistently improves performance, with the gains being more pronounced when MoE is applied to the decoder. By deploying MoE modules in both the encoder and decoder, R2E-IG achieves the best overall performance.

\subsubsection{Number of Experts}

We further investigate expert combinations by varying the expert cardinality. In R2E-IG, each routing step activates $k=3$ experts out of $m=8$, together with an always-on shared expert, and the expert intermediate dimension is set to 128. Removing the shared expert increases the average Gap from 1.1625\% to 1.2243\%, highlighting its role in capturing distribution-invariant knowledge. We additionally evaluate alternative configurations by adjusting the total number of experts $m$ and the number of activated experts $k$. We observe that, with a fixed $m$, increasing $k$ consistently reduces the Gap, indicating that activating more experts provides richer compositional capacity. Specifically, various experts can specialize in complementary distribution-specific patterns, and the enlarged set of expert combinations effectively increases the model capacity and expressiveness, leading to better solution quality. 

In addition, we observe a contrasting trend when varying $m$ under different sparsity levels $k$. When $k=3$, increasing $m$ slightly degrades performance, whereas for $k=1$ the performance improves as $m$ grows. A plausible explanation is the mismatch between activation budget $k$ and the size of the candidate expert pool $m$. With a very sparse budget ($k=1$), a larger pool may introduce higher routing uncertainty and amplify selection noise, making it harder for experts to specialize reliably. In contrast, with a less sparse budget ($k=3$), a larger pool provides richer complementary experts and more diverse expert combinations, thereby improving capacity and expressiveness.

\subsubsection{Dimension of Experts}

In the previous analysis related to number of experts, we fix the intermediate dimension of each expert to $IntDim=128$. In this case, the total intermediate budget can be computed as $IntDim(k+1)$, which scales with the number of activated experts $k$. Therefore, varying $k$ may change the model capacity, making the comparison less fair. To control for this factor, we additionally compare settings under a fixed total intermediate budget of 512. Concretely, we evaluate two representative configurations with the same budget: (i) $IntDim=128$, $k=3$ and (ii) $IntDim=256$, $k=1$ (both including the shared expert). When fixing $m$, increasing $k$ consistently yields better performance, suggesting that the number of available expert combinations is more critical than simply enlarging the intermediate dimension. When fixing $k$, we observe a similar trend as before, increasing $m$ can improve expressiveness and performance with a larger $k=3$. However, with a small $k=1$, enlarging $m$ may degrade the performance due to the mismatch between the limited activation budget and a large candidate expert pool. Overall, the results indicate that the combination of $m=8$ and $k=3$ provides the best trade-off under a constrained intermediate budget, which is the setting adopted throughout our experiments.


\section{Conclusion and Future Works}
\label{Conclusion}
\noindent Towards generalization-oriented models for VRPs, we propose R2E-IG, which brings the thoughts of MoE into improving cross-distribution generalization. Specifically, we design a novel R2E architecture that enhances expressiveness and improves robustness to unseen distributions compared with vanilla expert structure. We further develop an instance-level gating mechanism to achieve a favorable trade-off between solution quality and computational efficiency. Finally, we introduce a mixed-distribution training mechanism equipped with DWA, which adaptively emphasizes more informative data, stabilizes training, and improves overall training efficiency. Extensive experiments on both synthetic and real-world datasets demonstrate that R2E-IG consistently outperforms state-of-the-art baselines. While this work makes an initial step toward improving cross-distribution generalization with MoE-based modularization, several promising directions remain for future exploration: 1) \emph{scalability} to larger-scale VRPs by integrating R2E-IG with cross-scale generalization frameworks \cite{gao2024towards,wang2025distance,zhou2023towards}; 2) exploring the model’s \emph{performance ceiling} by enlarging the scale of model parameters; 3) developing a more \emph{fine-grained} training data design to further enhance generalization from the data perspective; 4) designing more \emph{efficient} and \emph{interpretable} gating mechanisms; and 5) constructing more \emph{flexible} modular architectures \cite{yang2020multi}.

\bibliographystyle{IEEEtran}
\bibliography{references.bib}

\end{document}